\documentclass{elsarticle}

\usepackage{hyperref}

\journal{XXX}

\usepackage{amssymb}
\usepackage{glossaries}
\usepackage{multirow}
\usepackage{graphicx}
\usepackage{lscape}
\usepackage{booktabs}
\usepackage{comment}
\usepackage{glossaries}
\usepackage{subfig}
\usepackage{url}

\begin{document}

\begin{frontmatter}

\newacronym{pe}{PE}{Private Equity}
\newacronym{vc}{VC}{Venture Capital}
\newacronym{ml}{ML}{Machine Learning}
\newacronym{ai}{AI}{Artificial Intelligence}
\newacronym{xgb}{XGB}{Extreme Gradient Boosting Model}
\newacronym{rf}{RF}{Random Forest}
\newacronym{svm}{SVM}{Support Vector Machine}
\newacronym{lr}{LR}{Linear Regression}
\newacronym{shap}{SHAP}{Shapley Additive Explanations}
\newacronym{smote}{SMOTE}{Synthetic Minority Oversampling Technique}
\newacronym{ch}{CH}{Companies House}

\title{A Data-Driven Framework for Identifying Investment Opportunities in Private Equity}





\author{Samantha Petersone$^1*$, Alwin Tan$^1*$, Richard Allmendinger$^1$, Sujit Roy$^1$, James Hales$^2$}
\address{ $^1$The University of Manchester, Booth St W, Manchester, United Kingdom, M15 6PB ; $^2$NorthEdge Capital LLP, St Pauls House, 23 Park Square S, Leeds LS1 2ND

$^*$ Equal Contribution \\
Email: richard.allmendinger@manchester.ac.uk }

\begin{abstract}
The core activity of a Private Equity (PE) firm is to invest into companies in order to provide the investors with profit, usually within 4-7 years. To invest into a company or not is typically done manually by looking at various performance indicators of the company and then making a decision often based on instinct. This process is rather unmanageable given the large number of companies to potentially invest. Moreover, as more data about company performance indicators becomes available and the number of different indicators one may want to consider increases, manual crawling and assessment of investment opportunities becomes inefficient and ultimately impossible. To address these issues, this paper proposes a framework for automated data-driven screening of investment opportunities and thus the recommendation of businesses to invest in. The framework draws on data from several sources to assess the financial and managerial position of a company, and then uses an explainable artificial intelligence (XAI) engine to suggest investment recommendations. The robustness of the model is validated using different AI algorithms, class imbalance-handling methods, and features extracted from the available data sources.
\end{abstract}


\end{frontmatter}


\section{Introduction}

Private equity (PE) is a broad alternative investment class \citep{fenn1997private}. A PE investment usually involves the purchase of part or all of a company that is not publicly traded on a stock exchange, even though investors may occasionally involve in the privatisation of publicly listed company. PE investments deal with companies from all stages. The funds raised from the investment are often tied to some purpose ranging from the product development for companies in their early stages to expansion in operations for mature companies. 


Over the past few decades, despite the rapid growth in the PE industry and the advancement in financial technology, studies focusing on the application of such techniques on investment decision support systems in the field of PE are still rare. This can be due to multiple reasons. Firstly, the \textbf{sparsity and scarcity of the historical deal data and financial data from small companies} has been considered as one of the major factors hampering advancements of data-driven methods in the PE market. According to the UK's audit exemption for small companies and micro entities\footnote{Further information can be found in Section 11 of Company Account Guidance}, these companies are not legally obligated to disclose their financial statement in full. They have the option to submit an abridged ('simpler') account or choose not to disclose their financial statement to the public depending on the region. In addition, PE firms are also not required to disclose the deal investment and exit values publicly, further aggravating the issue of data sparsity and scarcity.  This limits  further exploration using data-heavy Artificial Intelligence (AI) method, such as deep learning~\cite{goodfellow2016deep}. 


Next, the degree of complexity involved in applying computational methods in PE is exacerbated by the \textbf{qualitative and intuitive nature of investment decision making}. PE managers often report their decisions being based more on their `gut feeling' than hard-coded rules and data. An analysis has also been conducted on how the investment works based on entrepreneur profile or the idea~\cite{wang2020emphasizing}.  
The inability to quantify gut-feeling can also be associated with the important role played by the quality of the target companies management team and its relationships with the PE firm’s partners. Naturally, such relations are hard to quantify, contributing to the slow progress of AI adaptation in the PE industry. Moreover, black-box AI methods are of limited use in the PE domain because of the high risks involved in investments (primarily boiling down to the risk of losing a significant amount of money). This urges the need for explainable AI (XAI) methods~\cite{adadi2018peeking}.

Apart from this, the \textbf{holding period for PE investment is relatively long}, spanning a few to many years. Therefore, the estimation for any models using the exit outcome, for example, return on investment specific to the type of exit, may be less relevant as the market environment changes at a rapid pace. As a result, the model is less meaningful as they may not accurately address the current investment environment.

Finally, \textbf{data imbalance} has been repeatedly reported as an issue by many academic literature across different types of studies related to PE, as the number of no-deal companies significantly exceeds the number of deal companies. This imbalance in data create further challenges in making biased prediction. For example, when models are trained to predict investment into a company,  simply predicting no investment will be the best decision in most cases. Some studies have also used exit type data, which itself suffers from the survivorship bias and the under-representation of businesses going public after the deal compared to those who chose to stay private. Similarly, the fraction of businesses that have gone into administration is much smaller than that of active companies~\cite{bhat2011predicting}.

Having these data and application-specific challenges in mind, this study proposes and validates a framework for a data-driven tool to facilitate investment decisions. More specifically, the contributions of this paper include: 
\begin{itemize}
\item Motivation and framing the task of whether a PE firm should invest into a business or not. 

\item Proposal of a data-driven framework to automate the process of asset screening and thus the recommendation of businesses to invest or not. The framework combines data from several sources to ensure decisions are made holistically considering data related to private equity deals (Unquote data), finances (Fame) and management (Companies House). 
\item An exploratory data analysis (EDA) of the available data to gain a better understanding about the landscape of PE activities.

\item Validation of the data-driven framework for UK-registered businesses by benchmarking its performance for different (i)~prediction methods in terms of prediction accuracy and their ability to provide explainable investment recommendations, (ii)~class imbalance-handling methods, and (iii)~feature sets. 

\end{itemize}

The next section provides more background on how PE investment decisions are done and commonly considered decision criteria, followed by existing research on the application of AI in PE. Section~\ref{Metho} introduces the data-driven framework including an introduction to the available data, data-processing, and the AI methods considered. The EDA of the considered data is presented in Section~\ref{eda_marker}, followed by a validation of the framework in Section~\ref{resANDana}. Finally, Section~\ref{conclusions} concludes the study and discusses future work. 

\section{Literature Review}
Surprisingly, there is little existing research investigating the application of AI to support PE investment decisions. A related and more widely studied domain is the investment in stocks~\cite{ferreira2021artificial, gao2019empirical}. Although we can learn from that domain when it comes to optimizing PE investment decisions (e.g. in terms of investment criteria), the two types of investment are different and require different modeling techniques.

\subsection{Characteristics of good PE investment candidates}

A large number of investment criteria could be considered when making a PE investment decision. Which criteria should be considered depends amongst others on the drivers, type of an investment, and the focus of the particular PE business carrying out the investment. We will discuss these considerations and various investment criteria studied in the literature.

Arguably, the most important quantitative criteria are size and profitability-related. This includes Earning Before Interest, Taxes, Depreciation, and Amortization (EBITDA), cash flow and turnover of a business. There are also various qualitative criteria that have been considered when making an investment, such as the company's reputation, the competency of its management team~\cite{shepherd1999conjoint,clarysse2005early,broere2013decision}, market opportunity, and the product itself. Some of the more popular features from the qualitative criteria include the potential for high profit and exit, the risk of investment and the opportunity for the investors to involve in the prospects of the business~\citep{feeney1999private}.

A more recent study by Block et al.~\cite{block2019private} found the main criteria for successful investment to be revenue growth, value added of the product or service, management team's track record, international scalability, profitability and the business model. The results also indicated a strong preference for investments in company with reputable investors. This is supported by Dixon and Chong's study~\cite{dixon2014bayesian}, using the `investor rank' algorithm to detect successful early investors. They concluded that current investors in the company are a good indication for the future funding of the target firm. Similarly, Bhat and Zaelit~\cite{bhat2011predicting} find the first three funding rounds provide sufficient information to later stage investors on the probability of exit with a significant degree of confidence. 

Of the financial criteria, positive cash flow~\cite{block2019private} and EBITDA are often associated with the highest importance. EBITDA plays an important role in company pricing, as the price for the target firm is determined using a multiple of EBITDA, as long as the EBITDA satisfies some arbitrary minimum requirement~\cite{stafford2017replicating}. Previous studies show that the average enterprise value of a  target company is estimated to be around 7 times of the EBITDA (between 2.7 and 27 times)~\cite{ljungqvist_richardson_2003}. This multiple of EBITDA varies based on the firm's and fund's type, and the target company. For example, the EBITDA multiple for transactions involving smaller companies tends to be smaller than the larger transactions~\cite{puche2018deal}. 

Apart from the financial criteria, empirical literature also states that the calibre of the management has significant influence in the investment decision. Additionally, it has been observed investors' perceptions on industry selection is influenced by market conditions and global financial situation based on company stocks~\cite{dincer2016fuzzy}. Gottschlich et al. proposed a decision support system design that uses the crowd's recommendations and investors can use that in their investment decisions and further use it to manage a portfolio~\cite{gottschlich2014decision}. One of the earliest studies of the underlying drivers for success is the survey to identify common criteria selected by the PE firms~\cite{macmillan1985criteria}. The survey results were explained by the authors using an analogy of horse racing, where the market, company, and management are represented by the horse race, horse, and jockey respectively. According to the survey, 10 factors were identified as the indicator for success, covering everything from management and founder team to products, odds, and market. One interesting observation from these results was five of the top ten criteria were related to the management's experience or personality. In other words, the investors should first identify if the `jockey is fit to ride' - where the track record of the management is assessed to prove the team's ability to react to risk, and familiarity with the target market or industry. 

A more recent study conducted a meta analysis using 31 publications related to the drivers for success in the PE industry~\cite{song2008success}. Among the 31 studies, 24 factors were commonly identified across the literature. However, only 8 of these factors were statistically significant and positively correlated to investment performance. These 8 factors were the supply chain integration, market scope, firm age, size of the founding team, financial resources, management experience, founder experience, and the existence of patent protection. On the other hand, 5 of the 24 factors were identified as non-significant predictors. Counter-intuitively, research and development (R\&D) cost and competition intensity of the targeted company were part of these 5 non-significant factors. 

Another criteria mentioned in the empirical literature is the geographical location of the investing and target firms. In~\cite{zhong2018startup}, the authors find that PE investors have strong preference for local businesses indicated by the strong negative correlation between the probability of investment and the distance between the target firm and PE firm. This result may be due to the importance of professional networks in PE, and the close proximity between the managers of both firms, which simplifies communication. These findings are also supported by~\cite{florida1993venture} and~\cite{powell2002spatial}.

\subsection{Data-driven methods in PE}
In practice, the number of potential companies to invest in is often vast. For instance, according to Companies House, UK's registrar of companies, there are over 4 million private limited companies in the UK~\citep{ch_active_bus_stats}, hence it is not feasible for PE firms to manually screen all of these companies, even after filtering for key metrics, such as EBITDA and location. Automating part of the screening process could significantly improve the efficiency by showing only relevant companies to deal managers and may lead to more promising investment opportunities, which could have otherwise been overlooked by a human analyst. A survey conducted by von dem Knesebeck~\cite{bluefuture} reveals that improving the ability to find and execute deals is one of the main motivating factors for 59\% of the surveyed VC firms for improving their technology stack. This section provides an overview of the current developments in the PE space.

Dixon and Chong~\cite{dixon2014bayesian} examined the probability of successful investment focusing on private clean-tech businesses.\footnote{A successful company is defined as one having reached or filed for an initial public offering (IPO) or acquired by another investment firm at a minimum of 1.5 times the investment.} This was done using support vector machine (\acrshort{svm}) based on financial and other qualitative features (reputation of the company), and rank them using Bayesian methods. While this may be a novel approach to the problem, the authors did not provide sufficient evidence on the model performance and feedback.

Bhat and Zaelit~\cite{bhat2011predicting} apply random forests to predict the probability of PE investment to exit, using data on investor network and funding rounds. Companies are classified into bankrupt or non-bankrupt, and private or public after the investment. The model successfully predicts the probability of exit using only the funding information from the first three raises with an overall accuracy of 75\% and 82\% AUC. However, the model performance varies across sectors with the energy sector performing the best in the bankruptcy classification model, and bio-technologies and pharmaceutical sectors performing the best in the private or public company classification model. The authors also apply network analysis to the investor data and observed a positive correlation between the degree of centrality and the probability of exit.

The venture capital (\acrshort{vc}) -- a sub-field of the PE industry focused on supporting small businesses in their startup phase -- has been studied more widely than other fields in PE. Although many of the decision criteria for \acrshort{pe} and \acrshort{vc} differ, many of the challenges regarding computational approaches are similar. For instance, Zhong et al.~\cite{zhong2018startup} propose a probabilistic latent factor model to support investment decisions in the context of startups. The model is further extended with modern portfolio theory~\cite{elton2009modern} to provide recommendations on a personalised portfolio strategy. The reason for using a Bayesian probabilistic method is to overcome the data scarcity and sparsity issue present in this application. The authors predicts the potential returns based on features related to target company's geographical location, number and types of acquisitions and funding rounds, features of the company founders and the product and frequency of news on public media. The experimental study comprised several models and concluded that the proposed probabilistic model achieves a good performance in terms of accuracy; however, more importantly, the study concluded that investors may prefer to invest in a company with more competitors as this may be a sign of booming market.

In general, PE firms are starting to recognise the importance of intelligent data-driven decision making tools. This can be observed via the large investment in human capital (the surge in hiring for data scientist or machine learning engineers) and the rise of FinTech within the PE industry. For example, AlphaSense, an AI-based company that provides investment recommendations derived using natural language processing (NLP) and statistical modelling, or Two Six Capital that perform transaction due diligence using big data and machine learning algorithms. Both of these companies claim to be involved in significant transactions but the actual performance of their solutions are not disclosed. 

\begin{figure}[t!]
\includegraphics[width=\linewidth]{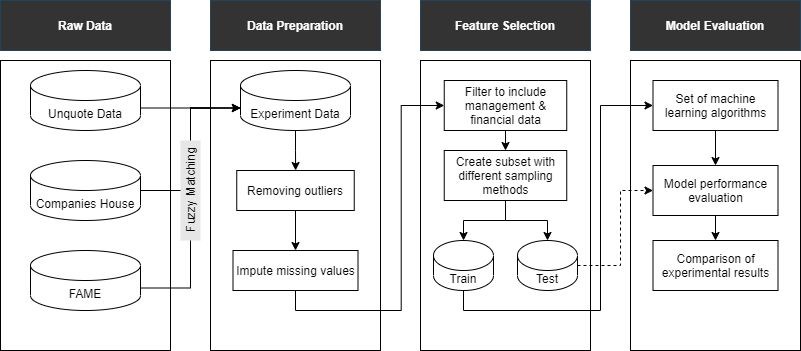}
\centering
\caption{Flow diagram of the proposed data-driven methodology for identifying investment opportunities in PE.}
\label{fig:methods}
\end{figure}

\section{Methodology}\label{Metho}
Figure~\ref{fig:methods} presents the proposed data-driven framework for identifying investment opportunities in PE. In summary, data from multiple relevant sources is combined, then pre-processed to be ready for the selection of relevant features, and training and validation of predictive models.

The study uses data from three sources: Unquote, Fame databases and Companies House (CH), where the objective is  to make investment recommendations for the UK market. The data from all the sources are merged into a single dataset  corresponding to different features for further analysis, which contains data points on companies' financial position, characteristics of the management team and investment status. Firstly, the data on \acrshort{pe} deals (from Unquote) is merged with the rest of the observations using a fuzzy matching algorithm (discussed in Section~\ref{FuzzyM}). Secondly, new management related features are engineered from the \acrshort{ch} Officer Appointment data, guided by the previous research on the role of management in \acrshort{pe}. Finally, six state-of-the-art AI models -- Logistic Regression (LR) \citep{maalouf2014weighted, blanco2020machine}, Decision Tree (DT) \citep{quinlan1990decision, jiang2007survey, decisiontree_2018}, Random Forest (RF) \citep{breiman1996bagging, breiman2001random, roy2020assessing}, k-Nearest Neighbors (kNN) \citep{jiang2007survey, zhu2010missing, zhu2014sparse}, Support Vector Machine (SVM) \citep{noble2006support, fan2000selecting} and eXtreme Gradient Boosting (XG Boost) \citep{chen2016xgboost, nwanganga_chapple_2020} -- are trained and validated using $k$-fold cross-validation ($k$=10).

The data provided by Unquote has 18 features and 3248 data points corresponding to each feature for the last 10 years. Each row of the data represents a deal executed with the equity value between \textsterling5m to \textsterling100m. Some of the important features are described as shown in Table~\ref{table:datadescription}. An Exploratory Data Analysis (EDA) was performed using this data to identify the characteristic of the deals, summarising factors like deal size, industry, and distribution of deal over the past 10 years.

\begin{table}[t!]
    \centering
    \caption{Description of selected features from the Unquote dataset.}
    \begin{tabular}{l l l }
    \toprule
         Feature  & Data Type & Example\\
          \midrule
        Deal Name  & String  & Distology \\
        Deal Date & Date & 2017-12-03 \\
        Country  & String & United Kingdom \\
        Deal Value (\textsterling m) & String & n/d(50-100m)\\
        & Numerical & 16.8 \\
        Industry   & String & Financial \\
        Business Description  & String &  Operator of restaurants\\
        Equity Lead  & String & NorthEdge Capital \\
        Region  & String & Eastern\\
        City & String & London\\
        \bottomrule
    \end{tabular}
    \label{table:datadescription}
\end{table}

\subsection{Fuzzy matching}\label{FuzzyM}
Having decided on set of relevant data sources, the next step is to link them. This can be challenging. In this study, it was not straightforward to link the Unquote data to the CH data as the deal data does not come with unique company identifiers (registration numbers) or standardised company names. For example, `Cera Ltd' may sometimes be recorded differently as `Cera', `Cera - Limited', `Cera Limited' or `Cera Holdings'. Due to lack of unique identifiers, the data was ambiguous in nature. Therefore, fuzzy matching, a technique designed to link string records when two strings are not identical, was implemented.

The Levenshtein distance, which computes the similarity of strings based on how many edits are required to make them identical, is the foundation of many string-based comparison algorithms. More precisely, the Levenshtein distance is the minimum number of operations that is needed to make two strings identical, with the operations being insertion, deletion or substitution of a character~\cite{levenshtein1966binary}. More formally, the distance between two strings $a$ and $b$ of length $|a|$ and $|b|$, respectively, is computed as 

\begin{equation}\operatorname{lev}_{a, b}(i, j)=\left\{\begin{array}{ll}
\max (i, j) \qquad\qquad\qquad\qquad\qquad\qquad\;\;\text { if } \min (i, j)=0 & \\
\min \left\{\begin{array}{ll}
\operatorname{lev}_{a, b}(i-1, j)+1 & \\
\operatorname{lev}_{a, b}(i, j-1)+1 & \text { otherwise, } \\
\operatorname{lev}_{a, b}(i-1, j-1)+1_{\left(a_{i} \neq b_{j}\right)} &\\
\end{array}\right.
\end{array}\right.\end{equation}
where $i \in \{ 1, 2, ..., |a|\} $ and $j \in \{ 1, 2, ..., |b|\}$ are indices of the strings $a$ and $b$, respectively, $\operatorname{lev}_{a, b}(i, j)$ is the distance between the first $i$ and $j$ characters in $a$ and $b$, and $1_{\left(a_{i} \neq b_{j}\right)}$ equals to 0 when the two strings are identical and 0 otherwise. The minimum function contains deletion, insertion and substitution elements. As the classic Levenshtein distance does not take into account the length of the strings, a normalized version Levenshtein ratio (LR)~\cite{marzal1993computation} is used instead.

\begin{table}[t!]
\centering
\caption{Fuzzy matching results. The number of companies in each fuzzy matching step. The full Unquote dataset information on each \acrshort{pe} deal. A number of companies have been involved in more than one \acrshort{pe} deal, thus the number of unique target companies is smaller than the number of records in the full dataset. Companies matched with their registration number with high enough confidence represents the number of unique companies that were matched with CH company name with at least a Levenshtein ratio of 90\% or with at least 70\% Levenshtein ratio and the same city as reported by the Unquote dataset. The Levenshtein ratio is computed after removing company legal status identifiers ('Limited', 'Ltd', 'Group', 'Holdings' etc) and converting all strings to lower case. Percentage represents the fraction of companies in the fuzzy matching step relative to the number of unique companies in the dataset.}
\resizebox{\columnwidth}{!}{%
\begin{tabular}{c c c} 
 \toprule
 Filtering & Number of observations & Percentage, \% \\ 
 \midrule
 Unquote dataset & 3248 & -\\ 
 Unique companies in the Unquote dataset & 2613 & 100\%\\
 $\geq90\%$ match confidence & 1529 & 59\%\\
 $\geq70\%$ and $<90\%$ match confidence & 585 & 22\%\\
 $\geq70\%$ and $<90\%$ match confidence, matched & 552 &21\%\\
 Companies matched  & 2081 & 80\%\\
 \bottomrule
\end{tabular}
}
\label{fuzzy_match}
\end{table}

The results of the matching are reported in Table~\ref{fuzzy_match}. Out of 2613 companies in the Unquote dataset, 2081 were matched adhering to the logic described, and 532 were discarded for various reasons including the incorrect labelling of company name, change in company name after transaction, and complication with the parent-subsidiary company structure.

\subsection{Data cleaning and engineering}
The management related features are derived from the Officer Appointment Snapshot with Resignations data provided by \acrshort{ch}. Company officer is a generic term for company directors and secretaries~\cite{gov_officers}. Thus, when referring to the company officer features, the term 'director features' and 'officer features' are used interchangeably for readability purposes. The following management features are derived: 

\begin{itemize}
\itemsep0em 
  \item Number of active directors
  \item Number of roles 
  \item Average experience at appointment
  \item Average number of previous companies 
  \item Average tenure prior to current role
  \item Cumulative experience at appointment
  \item Average age at appointment
\end{itemize}

A business can be incorporated by the founder or company formation agencies, which under the UK law are permitted to guide the process and submit incorporation documents, provide registered office address, help setting up a business bank account and provide ongoing company secretarial support~\cite{company_accounts_guidance_2020}. Thus formation agencies become officers in thousands of companies. For instance, two of the largest officers in the \acrshort{ch} Appointments Snapshot dataset, Temple Secretaries Limited and Company Directors Limited, are both linked to over 200,000 records each. This causes a large skew for the average number of previous companies and cumulative experience at appointment features, overshadowing the data on other director experiences. As the presence of formation agency in the company is not expected to provide any important information on the company for \acrshort{pe} investors, all institutional officers are removed from the dataset.

A number of the records are discarded from the dataset for the following reasons. Many of the companies received multiple investments in the same year. Because the financial data was joined by deal date, this leads to duplicate records, which would inflate the prediction results, so all duplicate company records are dropped. There are also holding companies, where only one of the child companies is active and the rest are dormant. As the dataset obtained from the Fame database contains both the parent and child businesses, and the financial data for those companies are very similar, one of the records is dropped to avoid information spill between the test and training sets in the modelling phase. 

Lastly, there is a risk of mismatches between the company names provided by the \acrshort{ch} and the Unquote database. In the cases where companies are matched incorrectly, the wrong financial and management data is joined with the company, which may result in large outliers. Thus, particular care is taken to ensure all erroneous matches are excluded from the final dataset. The top 2.5\% and bottom 2.5\% of the matched \acrshort{pe} and financial data, calculated by the features that contain the most prominent outliers, i.e. EBITDA, turnover and shareholder funds, are removed from the data set. Outliers in this case suggest that these records are matched with the wrong company or with a large parent company, so removing these records completely is preferable over opting for an outlier substitution method.

The final dataset contains 98385 records and 21 features, of which 814 companies have received \acrshort{pe} investment and 97571 companies have not received \acrshort{pe} investment. These firms will also be referred to as `deal companies' and `no-deal companies'. The final financial features include turnover, turnover growth, EBITDA, EBITDA margin, shareholder funds, number of employees, liquidity, return on shareholder equity (ROSE), profit margin, asset turnover, long term liabilities and minimum EBITDA and EBITDA margin from the past three years. The director features include number of active directors, number of roles in the company, average tenure at appointment, average director age at appointment, average experience and experience in the target company. The features also include the company age in years and a dummy feature, indicating whether a company has received a \acrshort{pe} deal previously or not. The summary statistics for all the features are reported in Section~\ref{eda_marker}.

\subsection{Modelling: Theoretical background}

Section~\ref{resANDana} will compare the ability of six AI models -- LR, DT, RF, kNN, SVM, XG Boost -- to predict the probability of a company receiving PE investment using financial and managerial features. The six models can be divided into two main categories using the complexity-explainability trade-offs. Three models, LR, DT, and kNN, emphasize on the ability to explain the reasoning behind the predicted probabilities to non-technical individuals but at a cost of predictive power potentially. The other three models, SVM, RF and XG Boost, have a higher predictive power but are less explainable.

\section{Exploratory Data Analysis (EDA)}
\label{eda_marker}
This section provides an EDA on key features of the (merged) data source to be used in Section~\ref{resANDana} for the experimental study.  

\begin{table}[t!]
    \centering
    \caption{Descriptive statistics for deal values.}
    \begin{tabular}{c c c c c c c c }
    \toprule
        Statistic & Min & 1st Qu. & Median & Mean & 3rd Qu. & Max & NAs\\
        \midrule
        Value & 5.000 & 7.765 & 14.000 & 23.462 & 30.000 & 100.000 & 1506\\
        \bottomrule
    \end{tabular}
    \label{tab:dealvalue}
\end{table}

\subsection{Deal value}
Table~\ref{tab:dealvalue} summarises the statistics of deal values. It can be seen that deals with lower equity values were significantly more likely to be executed, with the highest frequency between \pounds10 - \pounds15m. On top of that, a constant decrease in the number of deals can be observed as the deal value increases. Out of the 3200 deals, 1506 of them were recorded without disclosing the actual equity values, accounting for over 46\% of the data. The values of these deals were estimated using a conservative range by the industry experts. The distribution for these deals is summarised as shown in Table \ref{tab:nd_transaction}. A similar trend was observed in these deals even in the absence of equity values, where deal values and frequency of deals are inversely correlated.

\begin{table}[t]
    \centering
      \caption{Distribution of transaction with non-disclosed equity value.}
    \begin{tabular}{c c c  }
    \toprule
        Deal Value & Number of Transaction & Percentage\\
        \midrule
        n/d ($<$25\textsterling m) & 947 & 62.9\%\\
        n/d (25 - 50\textsterling m) & 344 & 22.8\%\\
        n/d (50 - 100\textsterling m) & 215 & 14.3\%\\
        \bottomrule
    \end{tabular}
    \label{tab:nd_transaction}
\end{table}

\subsection{Growth and cyclicality}
\begin{figure}[t]
    \centering
    \subfloat[\centering PE deals between 2010 and 2020.]{{\includegraphics[width=0.46\textwidth]{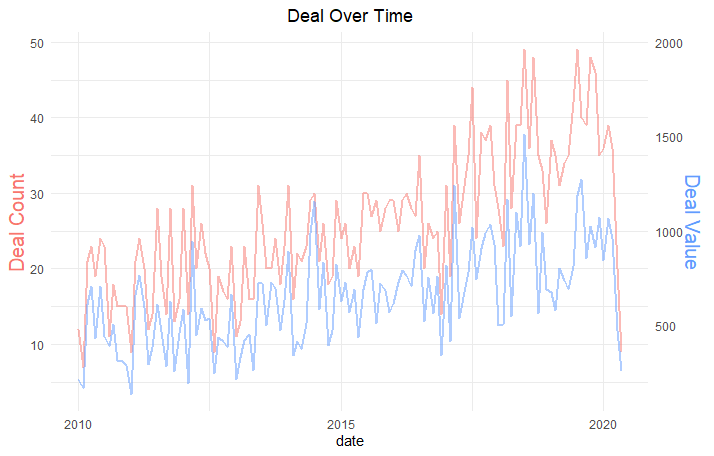} }}%
    \qquad
    \subfloat[\centering PE deals by month.]{{\includegraphics[width=0.46\textwidth]{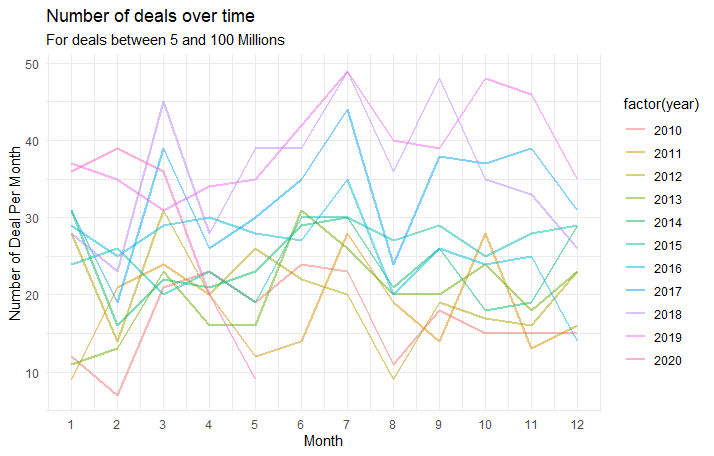} }}%
    \caption{PE growth and cyclicality.}%
    \label{fig:growth_cyclical}%
\end{figure}

Insights on the growth and cyclicality of the PE industry are summarised as shown in Figure~\ref{fig:growth_cyclical}. A clear upwards trend can be observed on the number of deals over time with some minor fluctuations in Figure \ref{fig:growth_cyclical}(a). On the other hand, constant spikes were observed during every March and July in the past 10 years as shown in Figure \ref{fig:growth_cyclical}(b), indicating higher activity levels for entry or exit.

\subsection{Company demographics}
\begin{figure}[t!]
    \centering
    \subfloat[\centering PE deals value by region.]{{\includegraphics[width=0.46\textwidth]{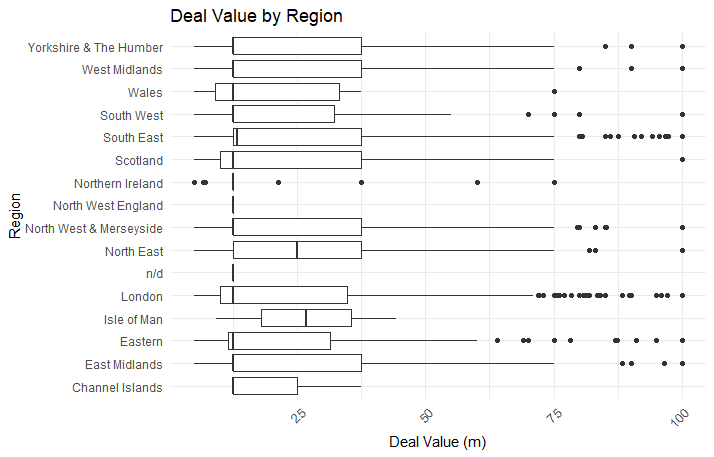} }}%
    \qquad
    \subfloat[\centering PE deals count by region.]{{\includegraphics[width=0.46\textwidth]{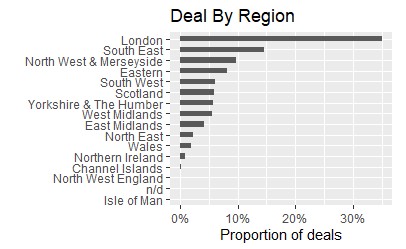} }}%
    \caption{PE deals by region.}%
    \label{fig:value_region}%
\end{figure}

The demographics of the companies that have received funding in the past including region and industry were explored as follows:

\vspace{+0.1cm}
\noindent \textbf{Region. }The distribution of companies receiving funding is summarised by location in Figure~\ref{fig:value_region}. Figure~\ref{fig:value_region}(a) shows that the distribution of deal values across regions was fairly consistent apart from some regions with limited observations, indicating that the PE firms do not necessarily pay a premium for the deal based on the location of the targeted company. However, 40\% of the deals in the past 10 years were executed on companies based in London followed by 15\% in the South East area of England. This may be due to the availability of resources as the business hub of the country. To note is that this is summary data across various PE companies; of course, on an individual basis, a PE firm may well limit itself to focus investments in certain regions and/or industries only.   

\begin{figure}[t]
    \centering
    \subfloat[\centering PE deals value by industry.]{{\includegraphics[width=0.46\textwidth]{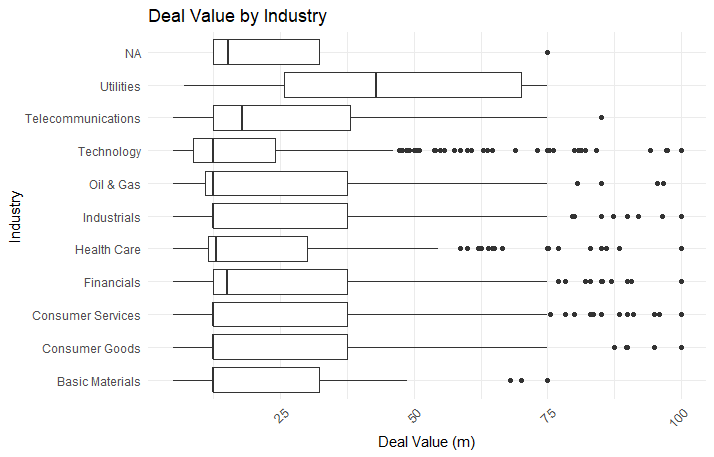} }}%
    \qquad
    \subfloat[\centering PE deals count by industry.]{{\includegraphics[width=0.46\textwidth]{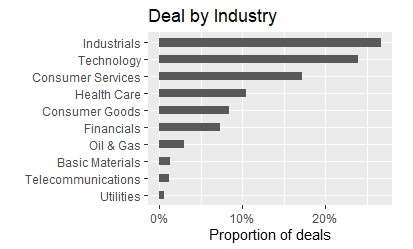} }}%
    \caption{PE deals by industry.}%
    \label{fig:value_industry}%
\end{figure}

\vspace{+0.1cm}
\noindent \textbf{Industry. }The distribution of companies by industry is summarised in Figure~\ref{fig:value_industry}. Figure~\ref{fig:value_industry}(a), shows that the funding received by companies from the technology and healthcare industry were slightly lower than the companies from other industry. However, despite the lower distribution in deal values, technology and healthcare industry were both observed with the highest number of outliers. The opposite can be observed for companies from the utility industry, as they demonstrated a distribution with higher deal values with no outliers. Next, in terms of the deal count, companies in the industrial sector will most likely receive funding, seconded by companies from the technology sector, which respectively makes up 27\% and 24\% of the deals between 2010 and 2020. 

\begin{figure}[t!]
\includegraphics[width=0.7\textwidth]{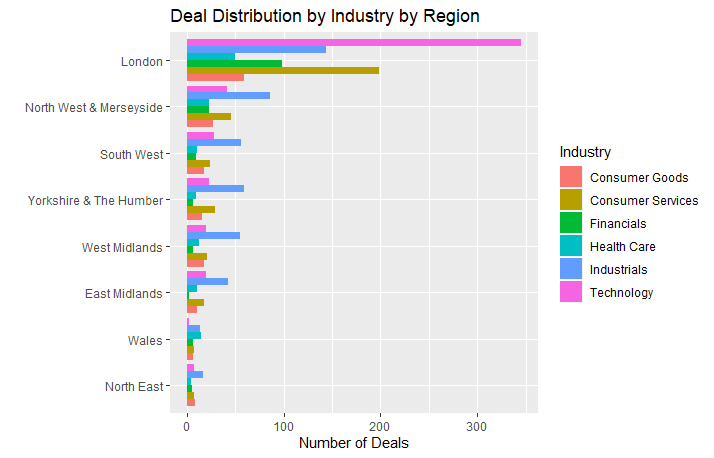}
\centering
\caption{PE deals by industry and regions.}
\label{fig:industry_region}
\end{figure}

\vspace{+0.1cm}
\noindent \textbf{Region and industry. }Some industries might be more prevalent in a specific region due to the availability of resources. Figure~\ref{fig:industry_region} shows the distribution of companies that received funding by industry and region across the UK. An interesting observation unique to deals in London is that the technology industry has the highest number of deals by a significant margin, seconded by the consumer services industry. On the other hand, for the remaining deals from other regions, the industrial sector consistently outnumbers the others, with less significant differences between the industries.

\subsection{Financials}
As stated previously, to reflect the financial and management features correctly, they are backdated to the time of investment as opposed to the current state. To ensure our findings are robust, we include confident matches (1796 companies) according to fuzzy matching in the EDA for financial and director data. Table~\ref{tab:stats} provides some statistics related to different financial features, which will be discussed next.

The average turnover for a company that has received funding is \textsterling40m. However, some companies reported extreme values. Examples of such outliers are The Currency Cloud Limited and The Sage Group PLC, who both recorded a turnover of over \textsterling2bn at the time of investment. On top of that, the average was also affected by the extreme from the opposite end, as one of the companies recorded a negative turnover of -\textsterling2.7m. Due to the lack of legal obligations on small companies as stated in the previous section, about 50\% of the matched companies had not filed their turnover at the time of the deal. 54\% of the companies recorded their EBITDA at the time of the deal, with an average EBITDA of \textsterling1.8m. However, more than 50\% of the companies reported a negative EBITDA. With regards to profit margin, the median and average values are only slightly in the positive with a reasonable number of companies having a negative profit margin. The majority of companies reported the shareholder fund at the time of deals, with less than 17\% in missing values. 57\% of companies recorded a shareholder fund of over \textsterling1m. 

The importance of these financial metrics will later be tested in the model building stage to determine if they should be included in the model.

\begin{table}[t]
\centering
\caption {Descriptive statistics for turnover, EBITDA, profit margin and shareholder fund.} \label{tab:stats}
\resizebox{1\textwidth}{!}{\begin{tabular}{cccccc}
\toprule
Parameter & Minimum & Median & Maximum & Average & \% of Missing Values    \\\midrule
Turnover (thousand) & -2,721 & 11,847 & 2,571,126 & 40,530 & 50.01\% \\ 
EBITDA (thousand) & -77,856 & 789 & 111,195 & 1,813 & 45.93\% \\ 
Profit margin & -75.75 & 4.25 & 57.11 & 2.68 & 57.52\% \\ 
Shareholder fund (thousand) & -187,719 & 1,528 & 1,577,433 & 8,166 & 16.81\% \\ \bottomrule
\end{tabular}}

\end{table}

\subsection{Management}
Although it is not mandatory for companies to report financial data, they are required by law to provide updated information for the appointment of directors and officers. For this reason, the analysis of management criteria was conducted with no missing values and the results are summarised in Table~\ref{tab:director}. 
\begin{table}[t]
    \centering
    \caption{Descriptive statistics for director data.}
    \resizebox{\columnwidth}{!}{%
    \begin{tabular}{ c  c  c  c  c  }
    \toprule
        Management criteria & Minimum & Median & Maximum & Average \\
        \midrule
        Directors & 1.00  & 4.00 & 134.00 & 4.47 \\
        Unique positions & 1.00  & 3.00 & 11.00 & 3.36 \\
        Corporate directors & 0.00  & 0.00 & 5.00 & 0.15 \\
        Directors with FTSE experience & 0.00  & 0.00 & 11.00 & 0.02 \\
        Directors with multiple positions & 0.00  & 3.00 & 22.00 & 2.91 \\
        Average length of tenancy & 0.00 & 3.57 & 28.00 & 5.00 \\
        Combined years of experience & 0.00  & 15.00 & 607.00 & 20.43 \\
        \bottomrule
    \end{tabular}
    }
    \label{tab:director}
\end{table}

Companies that have previously received funding have on average a smaller management team with fewer than five directors covering four different business functions. Moreover, companies tend to be filled with newer directors with an average tenancy duration of five years and 20 years in combined experience. Finally, despite the importance of having a director with experience from larger firms~\citep{macmillan1985criteria}, only 14 companies hired directors with previous experience in FTSE100 companies.

\section{Experimental Study}\label{resANDana}
Computational asset screening methods conventionally rely on financial and basic company information, such as company age, number of employees and industry. However, PE investors place as much of an importance on the company's management as on the company's financial performance. This poses challenges for the applications of traditional techniques on the \gls{pe} sector due to the difficulty of obtaining and quantifying information about the quality of company directors and management teams. We address these challenges by attempting to quantify a set of proxies for management team quality from officer appointment data, provided by the Companies House. We first conduct a comparative analysis of the performance of six AI models and three methods for dealing with class imbalance (Section~\ref{results_financial}). 
Hyperparameters of the models were set as suggested in the original papers. 

We then perform a more in depth analysis and contrast the predictive power of the company officer and company financial information features (Section~\ref{explain}). Generally AI models provide the prediction of the outcome variable. But we have also looked at the contribution of individual features/components in making that specific decision and tried to understand if it also correlate with general way of investing. We have used SHAP values~\cite{lundberg2017unified} to look at specific feature contribution in making a prediction in \gls{pe} investment decision making process. 
\subsection{Comparison of feature sets, algorithms and sampling techniques}\label{results_financial}

Table~\ref{tab:results_hold_out_alll} provides comparison of hold-out sample performance metrics for three different feature sets: financial and basic company information, company officer data, and the combination of both. Compared are the six AI models and three different techniques for dealing with class imbalance -- undersampling (at random from the majority class), oversampling with replacement (from the minority class) and SMOTE~\cite{chawla2002smote} -- to find interesting investment targets for \gls{pe} investors in a pool of companies. The table reports five different performance metrics: accuracy, precision, F1, ROC and recall rate. The training sample was balanced by either undersampling, oversampling with replacement or SMOTE, but the hold-out sample was left imbalanced with less than 1\% of the companies having received investment, to reflect the true state of the market. As it is important for the results in this study to both identify a large set of good investment opportunities, but also have relatively small number of false positives, the further analysis in the followings sections focuses on F1 score, which is the harmonic mean of recall rate and precision, and accounts for the class imbalance in the hold-out set.\\

\begin{table}[h]
\centering
\caption{Hold-out sample performance metrics by model, input features and sampling technique.}
\resizebox{\textwidth}{!}{%
\begin{tabular}{ccccccccccccccccc}
\toprule
Sampling & \multicolumn{1}{c}{\multirow{2}{*}{\textbf{Model}}} & \multicolumn{5}{c}{\textbf{Financial Features}} & \multicolumn{5}{c}{\textbf{Director Features}} & \multicolumn{5}{c}{\textbf{All Features}} \\ \cline{3-17} 
 technique & \multicolumn{1}{c}{} & \multicolumn{1}{c}{Accuracy} & \multicolumn{1}{c}{Precision} & \multicolumn{1}{c}{F1} & \multicolumn{1}{c}{ROC} & \multicolumn{1}{c}{Recall} & \multicolumn{1}{c}{Accuracy} & \multicolumn{1}{c}{Precision} & \multicolumn{1}{c}{F1} & \multicolumn{1}{c}{ROC} & \multicolumn{1}{c}{Recall} & \multicolumn{1}{c}{Accuracy} & \multicolumn{1}{c}{Precision} & \multicolumn{1}{c}{F1} & \multicolumn{1}{c}{ROC} & \multicolumn{1}{c}{Recall} \\ \midrule
\multirow{6}{*}{Undersampling} & LR & 0.72 & 0.019 & 0.036 & 0.69 & 0.66 & 0.62 & 0.014 & 0.027 & 0.64 & 0.66 & 0.71 & 0.019 & 0.038 & 0.71 & 0.71 \\
 & RF & \textbf{0.75} & \textbf{0.025} & \textbf{0.049} & \textbf{0.78} & \textbf{0.80} & \textbf{0.72} & \textbf{0.020} & \textbf{0.039} & 0.71 & 0.69 & 0.82 & 0.033 & 0.063 & 0.79 & 0.77 \\
 & XGBoost & 0.74 & 0.023 & 0.044 & 0.74 & 0.75 & 0.71 & \textbf{0.020} & \textbf{0.039} & \textbf{0.72} & 0.72 & 0.83 & \textbf{0.038} & \textbf{0.072} & \textbf{0.83} & \textbf{0.84} \\
 & SVM & 0.74 & 0.021 & 0.041 & 0.72 & 0.70 & 0.62 & 0.016 & 0.032 & 0.70 & \textbf{0.78} & \textbf{0.86} & 0.036 & 0.069 & 0.75 & 0.63 \\
 & KNN & 0.69 & 0.020 & 0.038 & 0.73 & 0.77 & 0.62 & 0.015 & 0.030 & 0.68 & 0.73 & 0.70 & 0.021 & 0.040 & 0.74 & 0.79 \\
 & DT & 0.70 & 0.018 & 0.035 & 0.69 & 0.67 & 0.66 & 0.016 & 0.032 & 0.68 & 0.70 & 0.75 & 0.023 & 0.045 & 0.74 & 0.73 \\ \hline
\multirow{6}{*}{Oversampling} & LR & 0.67 & 0.018 & 0.035 & \textbf{0.70} & 0.73 & 0.62 & 0.014 & 0.027 & 0.64 & 0.66 & 0.71 & 0.02 & 0.038 & \textbf{0.71} & 0.71 \\
 & RF & \textbf{0.99} & \textbf{0.455} & 0.059 & 0.52 & 0.03 & \textbf{0.99} & 0.000 & 0.000 & 0.50 & 0.00 & \textbf{0.99} & \textbf{0.36} & 0.058 & 0.52 & 0.03 \\
 & XGBoost & 0.97 & 0.084 & \textbf{0.128} & 0.62 & 0.27 & 0.93 & \textbf{0.031} & \textbf{0.056} & 0.60 & 0.26 & 0.98 & 0.18 & \textbf{0.241} & 0.67 & 0.35 \\
 & SVM & 0.80 & 0.024 & 0.045 & 0.69 & \textbf{0.58} & 0.73 & 0.021 & 0.042 & \textbf{0.73} & \textbf{0.72} & 0.94 & 0.07 & 0.130 & 0.76 & \textbf{0.58} \\
 & KNN & 0.97 & 0.046 & 0.070 & 0.56 & 0.14 & 0.97 & 0.021 & 0.031 & 0.52 & 0.06 & 0.98 & 0.05 & 0.068 & 0.55 & 0.11 \\
 & DT & 0.99 & 0.105 & 0.096 & 0.54 & 0.09 & 0.98 & 0.009 & 0.011 & 0.50 & 0.01 & 0.99 & 0.12 & 0.119 & 0.56 & 0.12 \\ \hline
\multirow{6}{*}{SMOTE} & LR & 0.68 & 0.017 & 0.034 & 0.69 & \textbf{0.70} & 0.63 & 0.014 & 0.027 & 0.64 & 0.65 & 0.70 & 0.02 & 0.037 & \textbf{0.70} & \textbf{0.71} \\
 & RF & 0.99 & 0.161 & 0.120 & 0.55 & 0.09 & 0.98 & 0.005 & 0.006 & 0.50 & 0.01 & 0.99 & 0.27 & 0.198 & 0.58 & 0.16 \\
 & XGBoost & \textbf{0.99} & \textbf{0.317} & \textbf{0.174} & 0.56 & 0.12 & \textbf{0.99} & \textbf{0.028} & 0.027 & 0.51 & 0.03 & \textbf{0.99} & \textbf{0.49} & \textbf{0.287} & 0.60 & 0.20 \\
 & SVM & 0.86 & 0.029 & 0.054 & \textbf{0.68} & 0.50 & 0.77 & 0.023 & \textbf{0.044} & \textbf{0.71} & \textbf{0.65} & 0.96 & 0.10 & 0.167 & 0.74 & 0.53 \\
 & KNN & 0.92 & 0.029 & 0.052 & 0.60 & 0.27 & 0.92 & 0.023 & 0.041 & 0.57 & 0.20 & 0.94 & 0.04 & 0.074 & 0.62 & 0.30 \\
 & DT & 0.98 & 0.061 & 0.082 & 0.56 & 0.13 & 0.97 & 0.009 & 0.012 & 0.50 & 0.02 & 0.97 & 0.08 & 0.112 & 0.60 & 0.22 \\ \bottomrule
\end{tabular}%
}

\label{tab:results_hold_out_alll}
\end{table}

Table~\ref{tab:results_hold_out_alll} show the performance of the six AI models, three imbalance-handling (sampling) methods, and three input feature sets using five metrics. Tables~\ref{tab:results_hold_out_alll} and~\ref{tab:avg_res_model_sample} show the average performance of each AI model by sampling techniques and the average performance of all models by input feature sets. It is obvious from the results that there is no clear winning setting. On average, models relying solely on financial features outperform those trained on director features only, across all sampling techniques. However, models that use both financial and director features consistently outperform those trained on one of the feature sets alone (Table \ref{tab:avg_res_sampl_features}). The average F1 score across all models for the financial data set, balanced by SMOTE, is 0.09, three times higher than that for director features (0.03). Using both feature sets yields an F1 score of 0.15, almost two times higher than that of financial features alone, and three times higher than the F1 score of director features alone. These results provide evidence that company's financial information alone is a significantly better predictor for the probability of investment than the information on company's officers, but both feature sets strongly compliment each other, and using both financial and management data on average yields almost two times better results.

\begin{table}[t]
\centering
\caption{Average hold-out sample performance metrics by the model and sampling technique. }
\resizebox{\textwidth}{!}{%
\begin{tabular}{cccccccccccccccc}
\toprule
 & \multicolumn{5}{c}{\textbf{Undersampling}} & \multicolumn{5}{c}{\textbf{Oversampling}} & \multicolumn{5}{c}{\textbf{SMOTE}} \\ \cline{2-16} 
\textbf{} & Accuracy & Precision & F1 & ROC & Recall & Accuracy & Precision & F1 & ROC & Recall & Accuracy & Precision & F1 & ROC & Recall \\ \midrule
LR & 0.682 & 0.017 & 0.034 & 0.680 & 0.677 & 0.667 & 0.017 & 0.033 & 0.684 & \textbf{0.700} & 0.668 & 0.017 & 0.032 & 0.677 & \textbf{0.686} \\
RF & \textbf{0.763} & 0.026 & 0.050 & 0.759 & 0.755 & \textbf{0.991} & \textbf{0.271} & 0.039 & 0.510 & 0.021 & 0.987 & 0.144 & 0.108 & 0.540 & 0.086 \\
XGB & 0.760 & \textbf{0.027} & \textbf{0.052} & \textbf{0.765} & \textbf{0.770} & 0.961 & 0.099 & \textbf{0.142} & 0.631 & 0.295 & \textbf{0.989} & \textbf{0.279} & \textbf{0.163} & 0.556 & 0.116 \\
SVM & 0.739 & 0.025 & 0.047 & 0.722 & 0.705 & 0.825 & 0.039 & 0.072 & \textbf{0.726} & 0.627 & 0.863 & 0.050 & 0.088 & \textbf{0.712} & 0.559 \\
KNN & 0.670 & 0.019 & 0.036 & 0.718 & 0.766 & 0.972 & 0.039 & 0.056 & 0.541 & 0.103 & 0.928 & 0.031 & 0.056 & 0.596 & 0.257 \\
DT & 0.703 & 0.019 & 0.037 & 0.701 & 0.698 & 0.985 & 0.078 & 0.075 & 0.533 & 0.074 & 0.975 & 0.048 & 0.069 & 0.551 & 0.120 \\ \bottomrule
\end{tabular}%
}
\label{tab:avg_res_model_sample}
\end{table}

\begin{table}[t]
\centering
\caption{ Average hold-out sample performance metrics by input features and sampling technique across all models}
\resizebox{\textwidth}{!}{%
\begin{tabular}{cccccccccccccccc}
\toprule
 & \multicolumn{5}{c}{\textbf{Undersampling}} & \multicolumn{5}{c}{\textbf{Oversampling}} & \multicolumn{5}{c}{\textbf{SMOTE}} \\ \cline{2-16} 
 & \multicolumn{1}{c}{\textbf{Accuracy}} & \multicolumn{1}{c}{\textbf{Precision}} & \multicolumn{1}{c}{\textbf{F1}} & \multicolumn{1}{c}{\textbf{ROC}} & \multicolumn{1}{c}{\textbf{Recall}} & \multicolumn{1}{c}{\textbf{Accuracy}} & \multicolumn{1}{c}{\textbf{Precision}} & \multicolumn{1}{c}{\textbf{F1}} & \multicolumn{1}{c}{\textbf{ROC}} & \multicolumn{1}{c}{\textbf{Recall}} & \multicolumn{1}{c}{\textbf{Accuracy}} & \multicolumn{1}{c}{\textbf{Precision}} & \multicolumn{1}{c}{\textbf{F1}} & \multicolumn{1}{c}{\textbf{ROC}} & \multicolumn{1}{c}{\textbf{Recall}} \\ \midrule
\textbf{Director} & 0.66 & 0.02 & 0.03 & 0.69 & 0.71 & 0.85 & 0.02 & 0.03 & 0.60 & 0.34 & 0.88 & 0.02 & 0.03 & 0.57 & 0.26 \\
\textbf{Financial} & 0.72 & 0.02 & 0.04 & 0.72 & 0.73 & 0.91 & 0.11 & 0.06 & 0.59 & 0.27 & 0.90 & 0.10 & 0.09 & 0.60 & 0.30 \\
\textbf{All} & 0.78 & 0.03 & 0.05 & 0.76 & 0.75 & 0.93 & 0.13 & 0.11 & 0.63 & 0.32 & 0.93 & 0.17 & 0.15 & 0.64 & 0.35 \\ \bottomrule
\end{tabular}%
}

\label{tab:avg_res_sampl_features}
\end{table}
There is also strong evidence for the benefits of oversampling methods over undersampling methods: models trained on SMOTE all-features dataset on average achieve three times higher F1 score (0.15) than that of under-sampled dataset (0.05), while oversampling with bagging achieves two times higher F1 score (0.11) relative to under-sampling.

Table \ref{tab:avg_res_model_sample} compares the predictive performance by model and sampling technique across all feature sets. Overall, the best results are achieved by tree-based algorithms in combination with SMOTE, followed by distance-based algorithms and, lastly, linear algorithms. XGBoost on average outperforms all other algorithms across all sampling techniques, achieving the highest average F1 score of 0.163. The results vary vastly across sampling techniques: for under-sampled dataset XGBoost provides a marginal improvement in the F1 score (0.052) of 0.02 points, compared to the next-best model, RF (F1 of 0.05). But for the SMOTE dataset, XGBoost provides an improvement in the average F1 score (0.16) by over 50\% compared to the next best model (RF, F1 OF 0.108). XGBoost is followed by RF with the average F1 score of 0.108 and SVM with the average F1 score of 0.088 (SMOTE sampling).

\begin{table}[t]
\centering
\caption{Logistic regression coefficients obtained by LR for all input features and SMOTE.}
\resizebox{\textwidth}{!}{%
\begin{tabular}{ccccc}
\toprule
\textbf{Feature} & \textbf{Coefficient} & \textbf{Standard Error} & \textbf{t-statistic} & \textbf{p-value} \\ \midrule
Constant & 4.720 & 0.09 & 53.99 & 0.00 \\
Company age, log & -1.773 & 0.02 & -111.68 & 0.00 \\
Turnover, log & 0.452 & 0.01 & 58.85 & 0.00 \\
Turnover growth & 0.002 & 0.00 & 13.12 & 0.00 \\
EBITDA & 0.023 & 0.01 & 3.87 & 0.00 \\
EBITDA\_margin & -0.029 & 0.00 & -23.57 & 0.00 \\
Shareholder funds & 0.000 & 0.00 & -32.68 & 0.00 \\
Employees & 0.001 & 0.00 & 18.74 & 0.00 \\
Liquidity, log & 0.046 & 0.01 & 4.62 & 0.00 \\
ROSE & 0.000 & 0.00 & -0.36 & 0.72 \\
Profit margin & 0.015 & 0.00 & 17.10 & 0.00 \\
Asset turnover, log & -0.024 & 0.00 & -25.13 & 0.00 \\
Long term liabilities & 0.001 & 0.00 & 5.76 & 0.00 \\
Min EBITDA & 0.279 & 0.01 & 22.13 & 0.00 \\
Min EBITDA margin & 0.000 & 0.00 & 0.41 & 0.69 \\
Number of active directors & -0.222 & 0.00 & -59.14 & 0.00 \\
Number of director roles, log & 1.585 & 0.02 & 101.20 & 0.00 \\
Average tenure & 0.251 & 0.00 & 77.37 & 0.00 \\
Average age at appointment, log & -0.100 & 0.00 & -58.47 & 0.00 \\
Number of previous companies, log & 0.140 & 0.01 & 20.22 & 0.00 \\
Experience in the company, log & 0.069 & 0.01 & 5.92 & 0.00 \\
Average experience at appointment, log & 0.394 & 0.01 & 35.92 & 0.00 \\
Number of directors with FTSE   experience  = 1 & 0.881 & 0.03 & 26.87 & 0.00 \\
Number of directors with FTSE experience   = 2 & -0.488 & 0.04 & -11.34 & 0.00 \\
Number of directors with FTSE   experience  \textgreater{}= 3 & 0.416 & 0.08 & 5.28 & 0.00 \\ \bottomrule
\end{tabular}%
}
\label{tab:LR_coeff}
\end{table}
The more granular model comparison by feature set and sampling technique in Table \ref{tab:results_hold_out_alll} provides additional insights about algorithms' ability to capture complex patterns and benefit from additional information. XGBoost achieves a high F1 score of 0.29 (all features, SMOTE balancing), but the performance of this algorithm is closely related to the amount and complexity of the data available. The capabilities of XGBoost are best seen for large datasets, i.e. over-sampled and using all features. XGBoost is an advanced algorithm, capable of capturing complex nonlinear relationships, but requires large amounts of data to be effective, and heavily overfits the smaller undersampled dataset. With increasing complexity, XGBoost performance improves significantly.  The findings are similar to other related studies.  For instance, \citep{munkhdalai_munkhdalai_namsrai_lee_ryu_2019} compare XGBoost, RF, SVM and LR on a credit scoring problem and find similar patterns in performance: XGBoost outperforms RF, SVM and LR in that order, measured by accuracy. But the authors found the difference between the best and worst performing algorithm less extreme: XGBoost provided an improved upon the accuracy between LR by only 6\%. We find there is very little change in the predictive power in LR, no matter the feature set or sampling technique. 

Table \ref{tab:LR_coeff} reports the LR model coefficients and test statistics. Almost all features are statistically significant with the exception of ROSE and minimum EBITDA margin. The variables with the highest economic significance include company age, EBITDA, turnover, min EBITDA and number of director roles. LR achieves an F1 score between 0.027 and 0.038 but the algorithm fails to capture complex patterns in the data set, and providing additional information does not improve the performance. Whereas for other models the predictive power increases with increasing complexity of sampling technique and larger data set. An exception, however, is the director feature data set. For models using only company officer information, the predictive power is relatively low across all sampling techniques with F1 score varying between 0 and 0.056.

\begin{figure}[t]
  \centering
  \includegraphics[width=\linewidth]{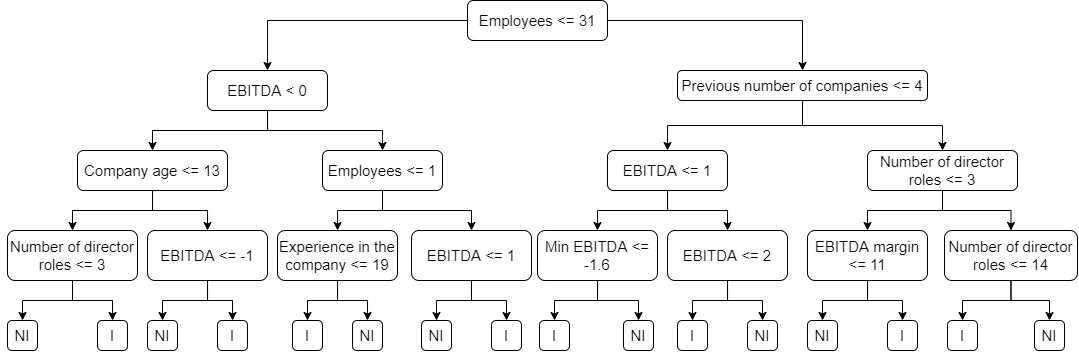}
  \caption{Decision tree obtained by DT for all input features and SMOTE.}
  \label{fig:decision_tree}
\end{figure}

DT achieves a slightly higher performance in terms of F1 (0.112), reporting similar features as the most important ones; this can be observed from the derived decision tree as shown in Figure~\ref{fig:decision_tree} using the combined input feature set. Among the first decision levels, there are EBITDA, number of previous companies, company age, number of employees and number of director roles. The decision tree also reveals several patterns that align with intuition. For example, the model suggests to invest in (i)~small, young companies with high number of directors and negative EBITDA, or  (ii)~companies with large number of employees with directors that have experience in a number of other companies and small EBITDA. If the directors do not have the aforementioned experience, higher number of director roles (i.e. breadth of experience  over depth of experience) are considered of similar importance.

Relative to XGBoost, non-boosted tree algorithms, i.e. RF and DT, the performance is more consistent: the models have less of a tendency to overfit for small datasets, but cannot capture as many complexities in the larger datasets as XGBoost. RF and SVM produce relatively consistent performance across different feature sets and sampling techniques, with relatively good results for both small and large data sets, and performance improvements when more information is provided. The F1 score of SVM varies between 0.032 and 0.167. RF performs well for small data sets, being the top algorithm for both under-sampled financial and director feature sets. An exception is the over sampled director feature data sets, where RF achieves a low F1 score of 0. Overall, our findings are consistent with previous literature. For instance, \citep{fernandez2014we} evaluate 170 classifiers from 17 algorithm families across 121 data sets; \gls{rf} was found to be the most likely best is classifier, achieving 90\% accuracy in 84\% of the datasets, followed by SVM, neural networks and boosting ensembles. In~\citep{huang_chen_wang_2007}, SVM, neural networks and DT classifiers are applied to a credit scoring problem and it has been found that SVM achieves identical classification accuracy as the other models but with fewer input features. It should, however, be noted that the runtime of SVM increases exponentially with the volume of input data, so in practice SVM models are not feasible for similar problems. We observed significantly longer training times for SVM for over-sampled data sets, relative to the other algorithms.



\subsection{Explainability}\label{explain}

Understanding the model decision process and the role of individual features is imperative for setting a good baseline benchmark for future \gls{pe} asset screening models and advancing the theoretical knowledge base of \gls{pe} decision making process. Examining certain feature interactions can be useful for bringing important details to the attention of investment analysts, when conducting the subsequent business analysis following the asset screening stage. This section is dedicated to the interpretation of the modelling results and the discussion on the importance of individual features and their interactions. In order to gain meaningful insights about the model decision process, it is crucial to have a good base model. As it was observed in Section~\ref{results_financial}, the XGBoost model trained on the SMOTE data set achieves the best predictive power out of all the algorithm combinations we studied. Therefore, the following section focuses on the analysis of this particular model only. We use \gls{shap} values to estimate feature importance and \gls{shap} interaction values to analyse the feature dependencies. 

\begin{figure}[t]
  \centering
  \includegraphics[width=\linewidth]{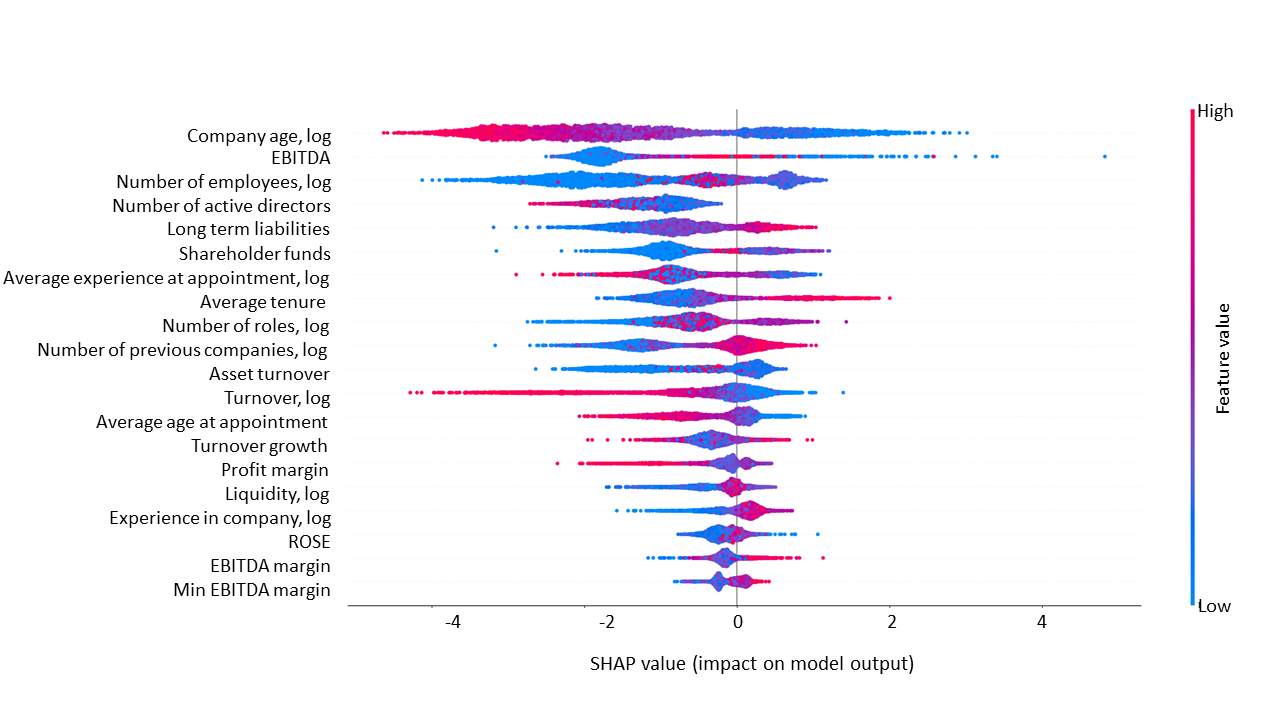}
  \caption{Feature importance with feature effects on model output, SMOTE}
  \label{fig:shap_importance_scatter}
\end{figure}

Figure \ref{fig:shap_importance_scatter} shows the top twenty features ordered by their importance in predicting the probability of receiving \gls{pe} investment in the hold-out sample, where importance is measured by the \gls{shap} values. The top five features are company age, EBITDA, the number of employees, the number of directors in the company and long term liabilities. The bottom features, that are not included in the plot, are number of directors with previous experience in a FTSE 500 company and the min EBITDA of the past three years. 
This reiterates the previous statement about \gls{pe} placing more importance on the management than the company's financial performance: while profitability margins are among the features with the lowest importance, management-related features, such as number of directors, average experience prior to the acquired company, the management team's average tenure and number of director roles have relatively high \gls{shap} values. The results described in the previous section demonstrate that financial and basic company information attain better model predictive power relative to management features alone, but according the \gls{shap} values, most management features carry higher importance in predicting the probability of the company receiving PE investment than the financial features. This discrepancy arises from feature interactions: management features alone are not strong predictors of probability for investment, but combined with financial information, the director features become essential.


\begin{figure}[t]
  \centering
  \includegraphics[width=\linewidth]{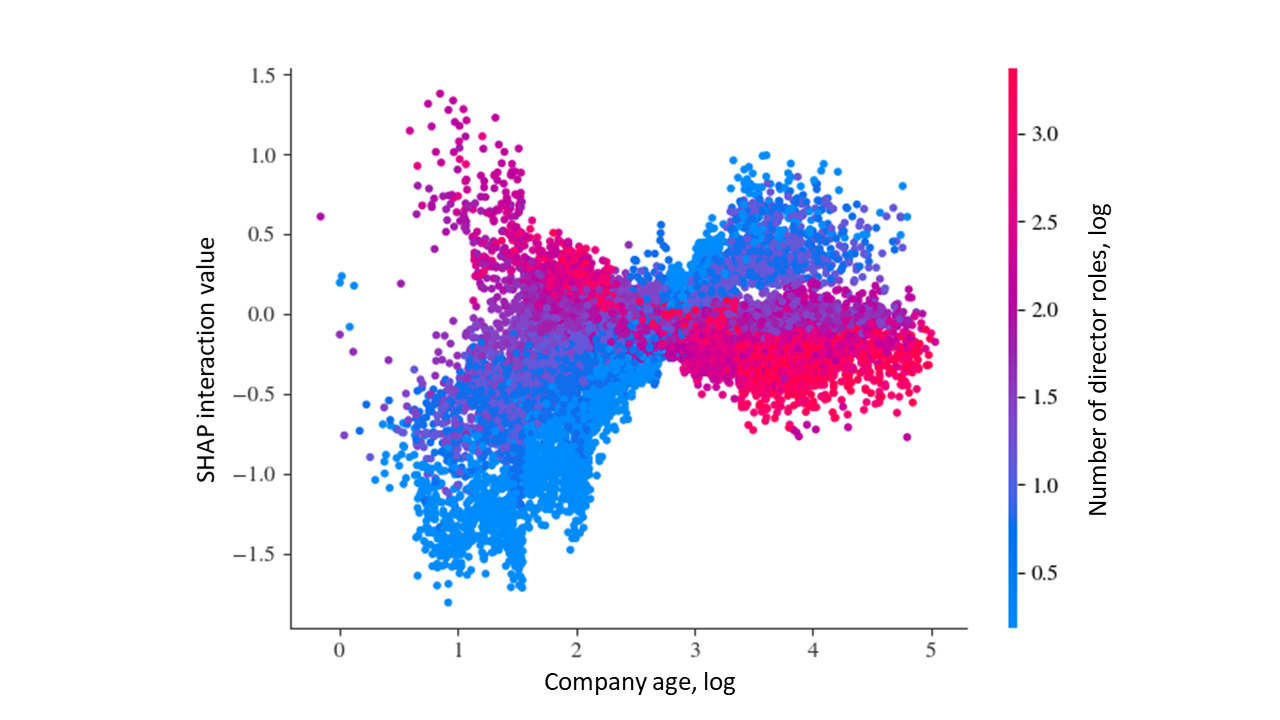}
  \caption{SHAP interaction values for company age vs number of director roles}
  \label{fig:comp_age_log_num_roles_log}
\end{figure}

\begin{figure}[h]
  \centering
  \includegraphics[width=\linewidth]{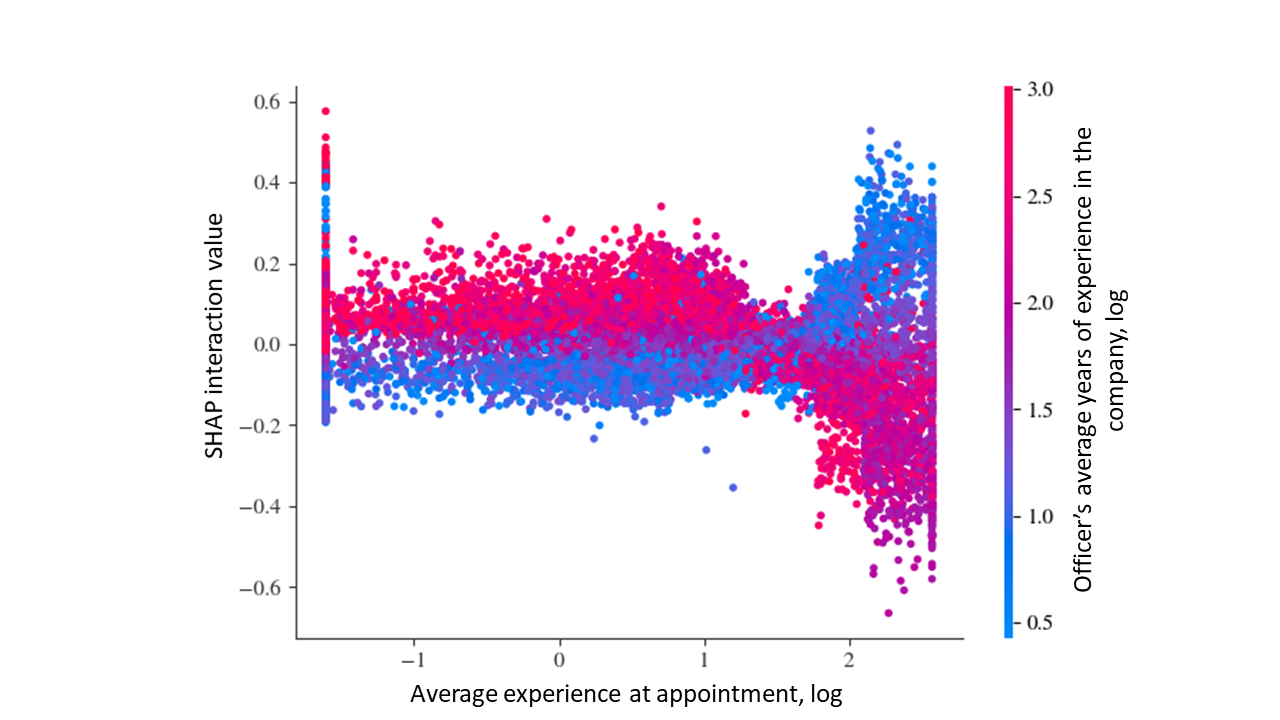}
  \caption{SHAP interaction values: average experience at the time of appointment and the average number of years in the company}
  \label{fig:avg_exp_appt_log_exp_in_comp_log}
\end{figure}

\begin{figure}[h]
  \centering
  \includegraphics[width=\linewidth]{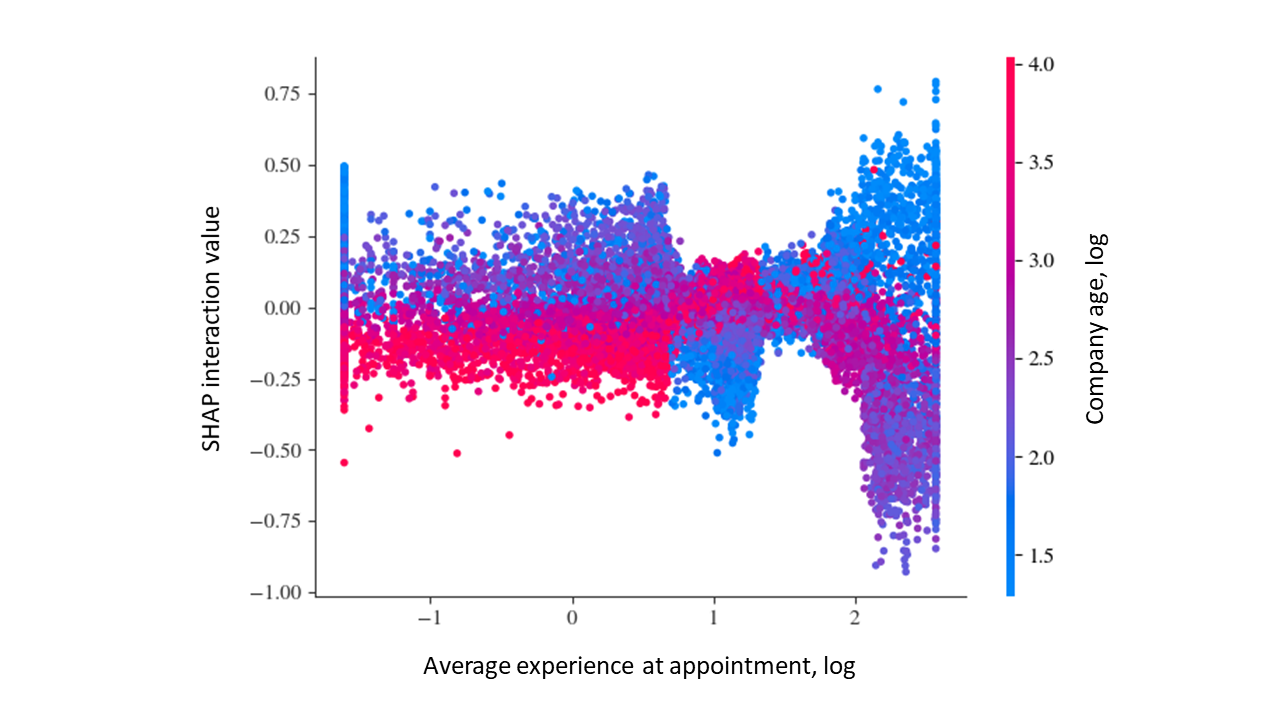}
  \caption{SHAP interaction values: average experience at appointment and the company age}
  \label{fig:avg_exp_appt_log_comp_age_log}
\end{figure}

Additionally, Figure \ref{fig:shap_importance_scatter} shows the relationships between the probabilities of receiving \gls{pe} investment and the feature values. The figure plots hold-out observations by feature and \gls{shap} value. The horizontal axis represents the \gls{shap} value of impact on model output for each feature shown on the y-axis, and the colour of each observation indicates the relative value of each observation, with red being high and blue low. Negative \gls{shap} values indicate a reduction in the probability of investment in the model decision making process, while positive \gls{shap} values increase it. Several features stand out with strong relations between their relative values and the impact on model output, particularly, company age, EBITDA, long term liabilities, average tenure, number of companies with experience as a director, turnover and the average age at appointment.

Company age (Figure~\ref{fig:comp_age_log_num_roles_log}), the feature with the highest impact on model output magnitude, has a strong negative correlation with the probability of receiving \gls{pe} investment. High age strongly impedes the chance of receiving investment, whereas young age increases it. In addition, according to \gls{shap}, the negative impact for age is stronger than the positive, respectively, company being relatively old is a better predictor of no \gls{pe} investment than young age being a predictor of investment. PE firms tend to invest in medium maturity companies with proven demand and product, so recently established businesses may carry more risk than desired, but older companies may be less appealing for several reasons, such as business model, lack of innovation, financial success and owner's attitude towards external investors. If a business is successful and shows potential signs of positive future returns, it is highly likely a \gls{pe} firm will have expressed the desire to invest in the company. In this case the business owners will either accept the offer (and the business would have been classified as an investment company) or the owners decline the offer due to the preference for full ownership or other reasons. Thus, the older the company gets, the smaller the probability for \gls{pe} investors to find `hidden gems', and the company's age becomes a predictor for investment by itself.

The financial features that proxy company size, i.e. EBITDA and turnover, appear to be more important than the relative profitability features, such as EBITDA margin or profit margin. This reveals the inherent flaws of AI models that rely solely on financial information: other than company size features, there is little information to be extracted from the profitability indicators, as \gls{pe} investors do not consider these as their primary point of interest. It must be noted, however, that this may not hold for leveraged buyout (LBO) firms and distressed company investments, where poor financial performance is the primary condition for investment and less importance is placed on the management team, as it is likely to be replaced after the investment. The EBITDA feature shows an interesting relationship between the feature variable and \gls{shap} values, that related to this observation. Observations with high EBITDA are concentrated around zero \gls{shap} value, while low values are more present at the extreme ends. The \gls{pe} firms concentrating on distressed businesses are likely to invest in low EBITDA companies, whereas firms looking for less risky investments are looking for higher EBITDAs.

Management features can be divided into three groups: experience (the average years of prior experience as a director, the average number of previous companies, average age at appointment, years of experience in the company and number of directors with experience in a FTSE 500 company), skill diversity (the number of active directors, number of different director roles) and turnover (tenure of the officer team members). According to Figure~\ref{fig:shap_importance_scatter}, the most important management features include features from each of the group, with the top three being the number of active directors, the average years of previous director experience as company officer, and the average director tenure in the team. There is a positive bias towards repeated founders as a company's success is closely related to the number of companies an owner has founded previously. The average number of companies the directors have held a management role at can be associated with more start-up experience and proven ability to start/scale up companies. The number of employees and number of directors are not correlated (0.08), meaning that higher number of directors in the company is not purely an implication of a larger company. High SHAP values are associated with low number or director roles and low number of employees or vice versa. 

The average tenure and the number of companies the directors had experience prior to being appointed have similar strong positive correlation with the probability of \gls{pe} investment. Average officer tenure in the company may be a proxy for the quality of the company's internal culture and founder's dedication to the company. High officer turnover may signal issues within the business. The number of roles within a company ranks the ninth,  showing that very high or low number of director roles in a company decrease the probability for investment, but the middle ranges increase it. This illustrates the significance of breadth of experience over multiple officers with similar experience.
  
For young to medium aged companies (approximately 3-8 years) it is preferable to have at least 4-5 directors, but for older companies above the age of 10 years, having no more than roughly 5 directors increases the probability for investment (Figure \ref{fig:comp_age_log_num_roles_log}). This can be explained with the breadth versus depth argument. For younger companies the inherent risk can be diversified with wider variety of skills, whereas for older companies when the risk is less substantial, it is more important to have a small team with extensive experience in the business and industry. A similar pattern can also be observed in Figure \ref{fig:avg_exp_appt_log_comp_age_log}, which shows the \gls{shap} interaction values between the company age and director experience. The probability of receiving \gls{pe} investment for a young company below the age of 5 years is significantly higher when the directors on average have had at least 5 years of experience prior to the appointment. For less experienced teams this probability is higher if the company is at least 7 years old. However, as discussed previously, companies aged above 20 years have negative \gls{shap} values associated with it, meaning companies above this age are less appealing investment targets. This can also be seen in Figure \ref{fig:avg_exp_appt_log_exp_in_comp_log}, which the \gls{shap} interaction values between the logged average years of experience at the time of appointment and the logged average experience in the company. According to the \gls{shap} values, the probability of \gls{pe} investment increases for companies with directors of at least 5 years of experience as officers in different companies prior to joining the target business and low average number of years in the target company. That is, \gls{pe} investors have a strong preference for young companies, founded by experienced founders. If the officers are less experienced, i.e. have between 0 and 5 years of experience as officers, the probability of investment increases if the officers have spent roughly 9 or more years in the business. First time founders may be riskier bet for \gls{pe} investors, thus they need to spend more time building the business and proving its success compared to experienced investors for \gls{pe}.

\section{Conclusion and Future Work}\label{conclusions}

This was the first study that investigated the application of AI to facilitate holistic \gls{pe} investment decision-making. After motivating the problem of investing into a business or not as a classification problem, we proposed a data-driven framework comprising the (fuzzy-matched) merger of 3 data sources (containing information about financial and managerial properties of businesses and previous PE deals), a data preparation pipeline, a feature selection and class imbalance method, and finally a model evaluation and explanation stage. Following an EDA of the available data to gain a better understanding and assess the application of AI, we conducted a comparative analysis on the model performance using three different feature sets across six different AI algorithms and three class imbalance-handling techniques. We then used SHAP values to examine the importance of financial and management features in the decision-making process and model predictive power. In summary, we found that:
\begin{itemize}
    \item Fundamental company and financial information has more predictive power of \gls{pe} investment than managerial data related to a company alone, but the best results are achieved using both feature sets.
  
    \item Shap values showed that company age has the highest impact on PE investment. High  age  strongly  impedes  the  chance  of  receiving investment, whereas young age increases it. The followed second important trait is EBITDA.
    
    \item The XGBoost model trained on the SMOTE data set achieves the best predictive power out of all the algorithm combinations studied.
    
    \item The best results are achieved by tree-based algorithms in combination with SMOTE, followed by distance-based algorithms and, lastly, linear algorithms.

\end{itemize}

This study can be extended in various ways. We validated the proposed framework for the United Kingdom but it would also be interesting to understand if and how drivers for investment decisions vary in other countries. In addition to management and financial data about companies, one can, of course, consider various other information to consider in investment decisions, such as ESG (Environmental, Social, and Governance) performance of a company. However, there is a lack of (open-source) data sources that contain such information (e.g. ESG) for all registered companies but it would be interesting to understand how this information relates to company valuation and PE investment decisions. Finally, while we considered binary decisions about investing/not investing in a company, it would be interesting to investigate whether a more fine-grained assessment (e.g. Likert scale for investment opportunities) leads to a more robust model; however, obtaining such labels for training is likely to be a time-consuming manual task.

\newpage
\appendix

\setcounter{table}{0}
\section{Results}
\begin{table}[htb]
\centering
\caption{Model evaluation metrics by model and input feature set, trained on under-sampled dataset. Train and test set panels represent the average 5-fold cross validation score. Training set is balanced by under-sampling the majority (investment) class. Hold-out panel represents the metric obtained by a model retrained on the training and test data combined.   }
\resizebox{\textwidth}{!}{%
\begin{tabular}{|cl|lllll|lllll|lllll|}
\hline
\multicolumn{1}{|l}{} &  & \multicolumn{5}{c|}{\textbf{Train}} & \multicolumn{5}{c|}{\textbf{Test}} & \multicolumn{5}{c|}{\textbf{Hold-out}} \\ \cline{3-17} 
\multicolumn{1}{|l}{} &  & Accuracy & Precision & F1 & ROC & Recall & Accuracy & Precision & F1 & ROC & Recall & Accuracy & Precision & F1 & ROC & Recall \\ \hline
\multicolumn{1}{|c|}{\multirow{6}{*}{\begin{tabular}[c]{@{}c@{}}Director \\ features\end{tabular}}} & LR & 0.66 & 0.65 & 0.67 & 0.66 & 0.70 & 0.62 & 0.015 & 0.029 & 0.66 & 0.69 & 0.62 & 0.014 & 0.027 & 0.64 & 0.66 \\
\multicolumn{1}{|c|}{} & RF & 1.00 & 1.00 & 1.00 & 1.00 & 1.00 & \textbf{0.71} & \textbf{0.021} & \textbf{0.042} & \textbf{0.75} & 0.78 & \textbf{0.72} & \textbf{0.020} & 0.039 & \textbf{0.71} & 0.69 \\
\multicolumn{1}{|c|}{} & XGB & \textbf{1.00} & \textbf{1.00} & \textbf{1.00} & \textbf{1.00} & \textbf{1.00} & 0.69 & 0.019 & 0.037 & 0.72 & 0.75 & 0.71 & 0.020 & \textbf{0.039} & 0.72 & 0.72 \\
\multicolumn{1}{|c|}{} & SVM & 0.74 & 0.70 & 0.76 & 0.74 & 0.83 & 0.61 & 0.017 & 0.032 & 0.71 & \textbf{0.81} & 0.62 & 0.016 & 0.032 & 0.70 & \textbf{0.78} \\
\multicolumn{1}{|c|}{} & KNN & 0.79 & 0.74 & 0.80 & 0.79 & 0.87 & 0.59 & 0.015 & 0.030 & 0.68 & 0.78 & 0.62 & 0.015 & 0.030 & 0.68 & 0.73 \\
\multicolumn{1}{|c|}{} & DT & 1.00 & 1.00 & 1.00 & 1.00 & 1.00 & 0.66 & 0.015 & 0.030 & 0.65 & 0.64 & 0.66 & 0.016 & 0.032 & 0.68 & 0.70 \\ \hline
\multicolumn{1}{|c|}{\multirow{6}{*}{\begin{tabular}[c]{@{}c@{}}Financial \\ features\end{tabular}}} & LR & 0.69 & 0.69 & 0.70 & 0.69 & 0.71 & 0.68 & 0.018 & 0.034 & 0.69 & 0.69 & 0.72 & 0.019 & 0.036 & 0.69 & 0.66 \\
\multicolumn{1}{|c|}{} & RF & 1.00 & 1.00 & 1.00 & 1.00 & 1.00 & \textbf{0.76} & \textbf{0.025} & \textbf{0.049} & \textbf{0.77} & \textbf{0.78} & \textbf{0.75} & \textbf{0.025} & \textbf{0.049} & \textbf{0.78} & \textbf{0.80} \\
\multicolumn{1}{|c|}{} & XGB & \textbf{1.00} & \textbf{1.00} & \textbf{1.00} & \textbf{1.00} & \textbf{1.00} & 0.74 & 0.024 & 0.046 & 0.75 & 0.76 & 0.74 & 0.023 & 0.044 & 0.74 & 0.75 \\
\multicolumn{1}{|c|}{} & SVM & 0.78 & 0.75 & 0.79 & 0.78 & 0.84 & 0.69 & 0.020 & 0.039 & 0.73 & 0.77 & 0.74 & 0.021 & 0.041 & 0.72 & 0.70 \\
\multicolumn{1}{|c|}{} & KNN & 0.81 & 0.79 & 0.82 & 0.81 & 0.85 & 0.69 & 0.019 & 0.037 & 0.71 & 0.74 & 0.69 & 0.020 & 0.038 & 0.73 & 0.77 \\
\multicolumn{1}{|c|}{} & DT & 1.00 & 1.00 & 1.00 & 1.00 & 1.00 & 0.70 & 0.017 & 0.034 & 0.68 & 0.66 & 0.70 & 0.018 & 0.035 & 0.69 & 0.67 \\ \hline
\multicolumn{1}{|c|}{\multirow{6}{*}{\begin{tabular}[c]{@{}c@{}}All \\ features\end{tabular}}} & LR & 0.73 & 0.72 & 0.73 & 0.73 & 0.74 & 0.70 & 0.019 & 0.037 & 0.71 & 0.72 & 0.71 & 0.019 & 0.038 & 0.71 & 0.71 \\
\multicolumn{1}{|c|}{} & RF & 1.00 & 1.00 & 1.00 & 1.00 & 1.00 & 0.81 & 0.035 & 0.068 & \textbf{0.83} & \textbf{0.84} & 0.82 & 0.033 & 0.063 & 0.79 & 0.77 \\
\multicolumn{1}{|c|}{} & XGB & \textbf{1.00} & \textbf{1.00} & \textbf{1.00} & \textbf{1.00} & \textbf{1.00} & \textbf{0.82} & \textbf{0.036} & \textbf{0.069} & 0.83 & 0.84 & 0.83 & \textbf{0.038} & \textbf{0.072} & \textbf{0.83} & \textbf{0.84} \\
\multicolumn{1}{|c|}{} & SVM & 0.83 & 0.80 & 0.83 & 0.83 & 0.87 & 0.79 & 0.029 & 0.055 & 0.78 & 0.77 & \textbf{0.86} & 0.036 & 0.069 & 0.75 & 0.63 \\
\multicolumn{1}{|c|}{} & KNN & 0.83 & 0.79 & 0.84 & 0.83 & 0.89 & 0.69 & 0.021 & 0.040 & 0.75 & 0.82 & 0.70 & 0.021 & 0.040 & 0.74 & 0.79 \\
\multicolumn{1}{|c|}{} & DT & 1.00 & 1.00 & 1.00 & 1.00 & 1.00 & 0.74 & 0.023 & 0.044 & 0.74 & 0.74 & 0.75 & 0.023 & 0.045 & 0.74 & 0.73 \\ \hline
\end{tabular}%
}

\label{tab:undersampling}
\end{table}

\begin{table}[h]
\centering
\caption{Model evaluation metrics by model and input feature set, SMOTE. Train and test set panels represent the average 5-fold cross validation score. Training set is balanced by over-sampling the minority class by SMOTE. Hold-out panel represents the metric obtained by a model retrained on the training and test data combined.   }
\resizebox{\textwidth}{!}{%
\begin{tabular}{|cl|lllll|lllll|lllll|}
\hline
\multicolumn{1}{|l}{} &  & \multicolumn{5}{c|}{\textbf{Train}} & \multicolumn{5}{c|}{\textbf{Test}} & \multicolumn{5}{c|}{\textbf{Hold-out}} \\ \cline{3-17} 
\multicolumn{1}{|l}{} &  & Accuracy & Precision & F1 & ROC & Recall & Accuracy & Precision & F1 & ROC & Recall & Accuracy & Precision & F1 & ROC & Recall \\ \hline
\multicolumn{1}{|c|}{\multirow{6}{*}{\begin{tabular}[c]{@{}c@{}}Director \\ features\end{tabular}}} & LR & 0.68 & 0.67 & 0.70 & 0.68 & 0.74 & 0.63 & 0.01 & 0.03 & 0.65 & 0.67 & 0.63 & 0.01 & 0.03 & 0.64 & 0.65 \\
\multicolumn{1}{|c|}{} & RF & 1.00 & 1.00 & 1.00 & 1.00 & 1.00 & \textbf{0.98} & 0.01 & 0.01 & 0.50 & 0.01 & 0.98 & 0.01 & 0.01 & 0.50 & 0.01 \\
\multicolumn{1}{|c|}{} & XGB & \textbf{1.00} & \textbf{1.00} & \textbf{1.00} & \textbf{1.00} & 0.99 & \textbf{0.98} & 0.00 & 0.00 & 0.50 & 0.00 & \textbf{0.99} & \textbf{0.03} & 0.03 & 0.51 & 0.03 \\
\multicolumn{1}{|c|}{} & SVM & 0.82 & 0.77 & 0.83 & 0.82 & 0.91 & 0.73 & \textbf{0.02} & \textbf{0.04} & \textbf{0.74} & \textbf{0.75} & 0.77 & 0.02 & \textbf{0.04} & \textbf{0.71} & \textbf{0.65} \\
\multicolumn{1}{|c|}{} & KNN & 0.97 & 0.94 & 0.97 & 0.97 & \textbf{1.00} & 0.92 & 0.02 & 0.04 & 0.56 & 0.20 & 0.92 & 0.02 & 0.04 & 0.57 & 0.20 \\
\multicolumn{1}{|c|}{} & DT & 1.00 & 1.00 & 1.00 & 1.00 & 1.00 & 0.97 & 0.01 & 0.01 & 0.50 & 0.02 & 0.97 & 0.01 & 0.01 & 0.50 & 0.02 \\ \hline
\multicolumn{1}{|c|}{\multirow{6}{*}{\begin{tabular}[c]{@{}c@{}}Financial \\ features\end{tabular}}} & LR & 0.71 & 0.70 & 0.72 & 0.71 & 0.75 & 0.67 & 0.02 & 0.03 & \textbf{0.70} & \textbf{0.73} & 0.68 & 0.02 & 0.03 & \textbf{0.69} & \textbf{0.70} \\
\multicolumn{1}{|c|}{} & RF & \textbf{1.00} & \textbf{1.00} & \textbf{1.00} & \textbf{1.00} & \textbf{1.00} & \textbf{0.99} & 0.19 & 0.14 & 0.55 & 0.11 & \textbf{0.99} & 0.16 & 0.12 & 0.55 & 0.09 \\
\multicolumn{1}{|c|}{} & XGB & \textbf{1.00} & \textbf{1.00} & \textbf{1.00} & \textbf{1.00} & \textbf{1.00} & \textbf{0.99} & \textbf{0.33} & \textbf{0.21} & 0.58 & 0.16 & \textbf{0.99} & \textbf{0.32} & \textbf{0.17} & 0.56 & 0.12 \\
\multicolumn{1}{|c|}{} & SVM & 0.86 & 0.82 & 0.87 & 0.86 & 0.92 & 0.83 & 0.02 & 0.05 & 0.66 & 0.50 & 0.86 & 0.03 & 0.05 & 0.68 & 0.50 \\
\multicolumn{1}{|c|}{} & KNN & 0.98 & 0.96 & 0.98 & 0.98 & 1.00 & 0.92 & 0.03 & 0.05 & 0.60 & 0.26 & 0.92 & 0.03 & 0.05 & 0.60 & 0.27 \\
\multicolumn{1}{|c|}{} & DT & 1.00 & 1.00 & 1.00 & 1.00 & 1.00 & 0.98 & 0.06 & 0.09 & 0.56 & 0.15 & 0.98 & 0.06 & 0.08 & 0.56 & 0.13 \\ \hline
\multicolumn{1}{|c|}{\multirow{6}{*}{\begin{tabular}[c]{@{}c@{}}All \\ features\end{tabular}}} & LR & 0.74 & 0.73 & 0.75 & 0.74 & 0.78 & 0.71 & 0.02 & 0.04 & 0.72 & \textbf{0.74} & 0.70 & 0.02 & 0.04 & \textbf{0.70} & \textbf{0.71} \\
\multicolumn{1}{|c|}{} & RF & \textbf{1.00} & \textbf{1.00} & \textbf{1.00} & \textbf{1.00} & \textbf{1.00} & 0.99 & 0.21 & 0.14 & 0.55 & 0.11 & 0.99 & 0.27 & 0.20 & 0.58 & 0.16 \\
\multicolumn{1}{|c|}{} & XGB & \textbf{1.00} & \textbf{1.00} & \textbf{1.00} & \textbf{1.00} & \textbf{1.00} & \textbf{0.99} & \textbf{0.53} & \textbf{0.28} & 0.59 & 0.19 & \textbf{0.99} & \textbf{0.49} & \textbf{0.29} & 0.60 & 0.20 \\
\multicolumn{1}{|c|}{} & SVM & 0.92 & 0.90 & 0.93 & 0.92 & 0.96 & 0.93 & 0.07 & 0.12 & \textbf{0.77} & 0.61 & 0.96 & 0.10 & 0.17 & 0.74 & 0.53 \\
\multicolumn{1}{|c|}{} & KNN & 0.98 & 0.95 & 0.98 & 0.98 & 1.00 & 0.93 & 0.05 & 0.08 & 0.66 & 0.38 & 0.94 & 0.04 & 0.07 & 0.62 & 0.30 \\
\multicolumn{1}{|c|}{} & DT & 1.00 & 1.00 & 1.00 & 1.00 & 1.00 & 0.97 & 0.06 & 0.09 & 0.58 & 0.18 & 0.97 & 0.08 & 0.11 & 0.60 & 0.22 \\ \hline
\end{tabular}%
}

\label{tab:smote}
\end{table}

\begin{table}[h]
\centering
\caption{The average model evaluation metrics by input feature set for under - sampled dataset. }
\resizebox{\textwidth}{!}{%
\begin{tabular}{|l|lllll|lllll|lllll|}
\hline
 & \multicolumn{5}{c|}{\textbf{Train}} & \multicolumn{5}{c|}{\textbf{Test}} & \multicolumn{5}{c|}{\textbf{Hold-out}} \\ \cline{2-16} 
 & Accuracy & Precision & F1 & ROC & Recall & Accuracy & Precision & F1 & ROC & Recall & Accuracy & Precision & F1 & ROC & Recall \\ \hline
Director & 0.86 & 0.85 & 0.87 & 0.86 & 0.90 & 0.65 & 0.02 & 0.03 & 0.69 & 0.74 & 0.66 & 0.02 & 0.03 & 0.69 & 0.71 \\
Financial & 0.88 & 0.87 & 0.88 & 0.88 & 0.90 & 0.71 & 0.02 & 0.04 & 0.72 & 0.73 & 0.72 & 0.02 & 0.04 & 0.72 & 0.73 \\
All & \textbf{0.90} & \textbf{0.89} & \textbf{0.90} & \textbf{0.90} & \textbf{0.92} & \textbf{0.76} & \textbf{0.03} & \textbf{0.05} & \textbf{0.77} & \textbf{0.79} & \textbf{0.78} & \textbf{0.03} & \textbf{0.05} & \textbf{0.76} & \textbf{0.75} \\ \hline
\end{tabular}%
}

\label{tab:undersample_by_input}
\end{table}

\begin{table}[h]
\centering
\caption{The average model evaluation metrics by input feature set for over - sampled (SMOTE) dataset. }
\resizebox{\textwidth}{!}{%
\begin{tabular}{|l|lllll|lllll|lllll|}
\hline
 & \multicolumn{5}{c|}{\textbf{Train}} & \multicolumn{5}{c|}{\textbf{Test}} & \multicolumn{5}{c|}{\textbf{Hold-out}} \\ \cline{2-16} 
 & Accuracy & \multicolumn{1}{l|}{Precision} & F1 & ROC & Recall & Accuracy & Precision & F1 & ROC & Recall & Accuracy & Precision & F1 & ROC & Recall \\ \hline
Director & 0.91 & 0.90 & 0.92 & 0.91 & 0.94 & 0.87 & 0.01 & 0.02 & 0.57 & 0.28 & 0.88 & 0.02 & 0.03 & 0.57 & 0.26 \\
Financial & 0.92 & 0.91 & 0.93 & 0.92 & 0.94 & 0.90 & 0.11 & 0.10 & 0.61 & 0.32 & 0.90 & 0.10 & 0.09 & 0.60 & 0.30 \\
All & \textbf{0.94} & \textbf{0.93} & \textbf{0.94} & \textbf{0.94} & \textbf{0.96} & \textbf{0.92} & \textbf{0.16} & \textbf{0.13} & \textbf{0.65} & \textbf{0.37} & \textbf{0.93} & \textbf{0.17} & \textbf{0.15} & \textbf{0.64} & \textbf{0.35} \\ \hline
\end{tabular}%
}

\label{tab:smote_by_input}
\end{table}

\begin{table}[h]
\centering
\caption{Average evaluation metrics by model for under - sampled dataset. }
\resizebox{\textwidth}{!}{%
\begin{tabular}{|l|lllll|lllll|lllll|}
\hline
 & \multicolumn{5}{c|}{\textbf{Train}} & \multicolumn{5}{c|}{\textbf{Test}} & \multicolumn{5}{c|}{\textbf{Hold - out}} \\ \cline{2-16} 
\textbf{} & Accuracy & Precision & F1 & ROC & Recall & Accuracy & Precision & F1 & ROC & Recall & Accuracy & Precision & F1 & ROC & Recall \\ \hline
LR & 0.69 & 0.69 & 0.70 & 0.69 & 0.72 & 0.67 & 0.02 & 0.03 & 0.69 & 0.70 & 0.68 & 0.02 & 0.03 & 0.68 & 0.68 \\
RF & 1.00 & 1.00 & 1.00 & 1.00 & 1.00 & \textbf{0.76} & \textbf{0.03} & \textbf{0.05} & \textbf{0.78} & \textbf{0.80} & \textbf{0.76} & 0.03 & 0.05 & 0.76 & 0.76 \\
XGB & \textbf{1.00} & \textbf{1.00} & \textbf{1.00} & \textbf{1.00} & \textbf{1.00} & 0.75 & 0.03 & 0.05 & 0.77 & 0.78 & 0.76 & \textbf{0.03} & \textbf{0.05} & \textbf{0.77} & \textbf{0.77} \\
SVM & 0.78 & 0.75 & 0.79 & 0.78 & 0.85 & 0.70 & 0.02 & 0.04 & 0.74 & 0.78 & 0.74 & 0.02 & 0.05 & 0.72 & 0.70 \\
KNN & 0.81 & 0.77 & 0.82 & 0.81 & 0.87 & 0.65 & 0.02 & 0.04 & 0.72 & 0.78 & 0.67 & 0.02 & 0.04 & 0.72 & 0.77 \\
DT & 1.00 & 1.00 & 1.00 & 1.00 & 1.00 & 0.70 & 0.02 & 0.04 & 0.69 & 0.68 & 0.70 & 0.02 & 0.04 & 0.70 & 0.70 \\ \hline
\end{tabular}%
}

\label{tab:undersampled_by_model}
\end{table}

\begin{table}[h]
\centering
\caption{Average evaluation metrics by model for SMOTE dataset}
\resizebox{\textwidth}{!}{%
\begin{tabular}{|l|lllll|lllll|lllll|}
\hline
 & \multicolumn{5}{c|}{\textbf{Train}} & \multicolumn{5}{c|}{\textbf{Test}} & \multicolumn{5}{c|}{\textbf{Hold - out}} \\ \cline{2-16} 
\textbf{} & Accuracy & Precision & F1 & ROC & Recall & Accuracy & Precision & F1 & ROC & Recall & Accuracy & Precision & F1 & ROC & Recall \\ \hline
LR & 0.71 & 0.70 & 0.72 & 0.71 & 0.75 & 0.67 & 0.02 & 0.03 & 0.69 & \textbf{0.71} & 0.67 & 0.02 & 0.03 & 0.68 & \textbf{0.69} \\
RF & \textbf{1.00} & \textbf{1.00} & \textbf{1.00} & \textbf{1.00} & \textbf{1.00} & 0.99 & 0.14 & 0.10 & 0.54 & 0.08 & 0.99 & 0.14 & 0.11 & 0.54 & 0.09 \\
XGB & \textbf{1.00} & \textbf{1.00} & \textbf{1.00} & \textbf{1.00} & \textbf{1.00} & \textbf{0.99} & \textbf{0.29} & \textbf{0.17} & 0.56 & 0.12 & \textbf{0.99} & \textbf{0.28} & \textbf{0.16} & 0.56 & 0.12 \\
SVM & 0.87 & 0.83 & 0.87 & 0.87 & 0.93 & 0.83 & 0.04 & 0.07 & \textbf{0.72} & 0.62 & 0.86 & 0.05 & 0.09 & \textbf{0.71} & 0.56 \\
KNN & 0.97 & 0.95 & 0.97 & 0.97 & \textbf{1.00} & 0.93 & 0.03 & 0.06 & 0.61 & 0.28 & 0.93 & 0.03 & 0.06 & 0.60 & 0.26 \\
DT & \textbf{1.00} & \textbf{1.00} & \textbf{1.00} & \textbf{1.00} & \textbf{1.00} & 0.97 & 0.04 & 0.06 & 0.55 & 0.11 & 0.97 & 0.05 & 0.07 & 0.55 & 0.12 \\ \hline
\end{tabular}%
}

\label{tab:smote_by_model}
\end{table}

\clearpage 
\bibliography{mybibfile}

\end{document}